\documentclass[lettersize,journal]{IEEEtran}
\usepackage{amsmath,amsfonts}
\usepackage{algorithmic}
\usepackage{array}
\usepackage{textcomp}
\usepackage{stfloats}
\usepackage{url}
\usepackage{verbatim}
\usepackage{graphicx}
\usepackage{subcaption}
\usepackage{caption}
\usepackage{booktabs}
\def\BibTeX{{\rm B\kern-.05em{\sc i\kern-.025em b}\kern-.08em
    T\kern-.1667em\lower.7ex\hbox{E}\kern-.125emX}}
\usepackage{balance}

\begin{document}
\title{Parameter-Efficient Domain Adaption for CSI Crowd-Counting via Self-Supervised Learning with Adapter Modules}
\author{Oliver Custance, Saad Khan, Simon Parkinson, Quan Z. Sheng
\thanks{Manuscript created September, 2025; This work was developed by the IEEE Publication Technology Department. This work is distributed under the \LaTeX \ Project Public License (LPPL) ( http://www.latex-project.org/ ) version 1.3. A copy of the LPPL, version 1.3, is included in the base \LaTeX \ documentation of all distributions of \LaTeX \ released 2003/12/01 or later. The opinions expressed here are entirely that of the author. No warranty is expressed or implied. User assumes all risk.}}

\markboth{Journal of \LaTeX\ Class Files,~Vol.~xx, No.~x, September~2025}%
{How to Use the IEEEtran \LaTeX \ Templates}

\maketitle

\begin{abstract}
Device-free crowd-counting using WiFi Channel State Information (CSI) is a key enabling technology for a new generation of privacy-preserving Internet of Things (IoT) applications. However, practical deployment is severely hampered by the domain shift problem, where models trained in one environment fail to generalise to another. To overcome this, we propose a novel two-stage framework centred on a CSI-ResNet-A architecture. This model is pre-trained via self-supervised contrastive learning to learn domain-invariant representations and leverages lightweight Adapter modules for highly efficient fine-tuning. The resulting event sequence is then processed by a stateful counting machine to produce a final, stable occupancy estimate. We validate our framework extensively. On our WiFlow dataset, our unsupervised approach excels in a 10-shot learning scenario, achieving a final Mean Absolute Error (MAE) of just 0.44--a task where supervised baselines fail. To formally quantify robustness, we introduce the Generalisation Index (GI), on which our model scores near-perfectly, confirming its ability to generalise. Furthermore, our framework sets a new state-of-the-art public WiAR benchmark with 98.8\% accuracy. Our ablation studies reveal the core strength of our design: adapter-based fine-tuning achieves performance within 1\% of a full fine-tune (98.84\% vs. 99.67\%) while training 97.2\% fewer parameters. Our work provides a practical and scalable solution for developing robust sensing systems ready for real-world IoT deployments.
\end{abstract}


\section{Introduction}
The proliferation of the Internet of Things (IoT) is fundamentally reshaping our interaction with the physical world, transforming static spaces into intelligent, adaptive ecosystems~\cite{kumar2019internet}. A critical capability underpinning this evolution is automated crowd-counting: the task of accurately and persistently estimating human occupancy within a defined area. This is not merely a data-gathering exercise; it is a foundational technology that drives tangible improvements across a spectrum of applications. Within smart building management, for instance, granular occupancy data enables dynamic, zone-specific control of HVAC and lighting systems, yielding substantial energy savings and aligning with pressing environmental sustainability goals~\cite{chaudhari2024fundamentals}. In the commercial sector, retailers can leverage real-time footfall analytics to optimise store layouts, manage staffing levels, and enhance customer experience~\cite{tlapana2021impact}. Moreover, in public safety, the ability to monitor crowd density is paramount to ensuring regulatory compliance, orchestrating efficient evacuations, and preventing dangerous overcrowding in transport hubs and event venues~\cite{liu2021emergency, weng2023review}.

For decades, the de facto solution for this task has been vision-based systems. Leveraging  progress in computer vision, modern approaches utilise sophisticated deep learning models like Convolutional Neural Networks (CNNs) for direct object detection or, more advanced, for regressing crowd density maps~\cite{ilyas2019convolutional}. Although these systems can achieve high accuracy under ideal conditions, their practical utility is severely undermined by a set of inherent and often intractable limitations. The most salient of these is the issue of privacy. The continuous visual surveillance required is often ethically contentious and legally problematic under data protection regulations such as GDPR, rendering cameras unsuitable for private or semi-private spaces such as corporate offices, healthcare facilities, and residential homes~\cite{mihaescu2024gdpr}. Operationally, their performance is notoriously brittle, degrading significantly in the presence of common real-world challenges such as poor illumination, glare, physical occlusions from furniture or other people, and adverse weather conditions~\cite{kristoffersen2016pedestrian}.

These fundamental drawbacks have catalysed a search for alternative sensing modalities that are inherently robust and privacy-preserving. WiFi-based device-free sensing has emerged as a uniquely powerful candidate, repurposing the ubiquitous wireless infrastructure already present in virtually all modern buildings~\cite{zhu2025experience}. This technology exploits the physical layer information from standard WiFi protocols, specifically the Channel State Information (CSI). CSI provides a fine-grained characterisation of the radio frequency (RF) channel, capturing the complex interplay of signal reflections, diffractions, and scattering that occurs as radio waves propagate from a transmitter to a receiver. The human body, which is composed largely of water, significantly absorbs and scatters RF signals. Consequently, a person's presence and movement create distinct, measurable perturbations in the CSI's amplitude and phase across its multiple subcarriers~\cite{al2019channel}. By analysing the temporal dynamics of these CSI streams, it is possible to infer a rich variety of human activities, from gestures to occupancy, without capturing any personally identifiable information.

However, despite its potential, the widespread real-world deployment of WiFi-based sensing has been critically impeded by a single formidable obstacle: the domain shift problem. The very sensitivity that makes CSI a rich data source also makes it exquisitely dependent on the specific physical environment. A machine learning model trained to map the CSI patterns to occupancy in one room (source domain) implicitly learns the unique RF ``fingerprint'' of that space, which is a complex function of its geometry, furniture and construction materials. When this model is deployed in a new location (the target domain), this learnt fingerprint becomes obsolete. The statistical relationship between CSI patterns and human activity is fundamentally altered, causing a catastrophic collapse in the predictive performance of the model~\cite{zhao2024knn}. This environmental overfitting is far more severe than in other fields; even minor changes, such as moving a metal cabinet, can create new signal reflection paths that corrupt the data distribution in ways that have no visual parallel. The conventional solution regarding collecting a large, manually labelled dataset for every new room is prohibitively expensive, labour intensive, and fundamentally unscalable for any practical IoT application. This has been the single greatest barrier preventing WiFi sensing from moving out of the lab and into the real world.

To definitively solve this challenge, we propose a novel framework for cross-domain crowd-counting designed explicitly for robustness, scalability, and data efficiency. Our approach is founded on the core hypothesis that while raw CSI signals are domain specific, the underlying latent representations of human-induced signal dynamics can be learnt in a domain-invariant manner. We leverage self-supervised contrastive learning on vast quantities of unlabelled CSI data to pre-train a deep encoder. This data was collected from 10 participants performing entry and exit sequences across two distinct indoor environments (laboratory and classroom), with trials conducted under varied occupancy scenarios involving groups of 2, 5, and 9 users to ensure a diverse set of signal perturbations. This process compels the model to discover the essential and transferable features of human movement by learning to distinguish between different types of CSI perturbations, rather than memorising the static RF background of a particular room. This robustly pre-trained model then serves as a powerful, general-purpose feature extractor.

Our framework operates as a two-stage ``classify-then-count'' pipeline. It first uses the adapted encoder to classify short, overlapping windows of CSI data into discrete motion events (Enter, Exit, or no event). This stream of classified events is then consumed by a deterministic, stateful counting machine that integrates these events over time to maintain a stable and accurate final occupancy count. This architectural decision to decouple the high-frequency perceptual task from the low-frequency state-tracking task makes the system more resilient to transient noise and inherently more interpretable than monolithic end-to-end regression models, which often produce physically impossible or volatile predictions.

The contributions in this paper are as follows:
\begin{enumerate}
\item{A Novel Two-Stage Framework for Robust Crowd-Counting. We separate the problem into a classification stage and a counting stage. This ``classify-then-count'' pipeline is inherently more robust and interpretable than direct regression approaches, using a stateful machine to integrate events over time and provide a stable occupancy estimate resilient to spurious signal fluctuations.}
\item{A Parameter-Efficient Architecture for Cross-Domain Adaptation. We propose CSI-ResNet-A, a deep residual network whose `ResidualBlock' units integrate a `SqueezeExcitation' module for channel-wise feature recalibration. Crucially, these blocks are augmented with lightweight Adapter modules, which employ a bottleneck architecture using 1D convolutions. During adaptation, the main encoder weights are frozen and only these small adapters are trained. This isolates domain-specific adjustments to a tiny fraction of the total model parameters, enabling rapid, data-efficient fine-tuning without catastrophic forgetting.}
\item{A Comprehensive Cross-Domain Evaluation Demonstrating State-of-the-Art Performance. We conducted extensive experiments on our proprietary WiFlow dataset and the public WiAR benchmark. Our framework significantly outperforms existing supervised methods and adversarial techniques in challenging few-shot and zero-shot scenarios, validating the efficacy of our approach.}
\item{The Introduction of a Novel Metric for Quantifying Generalisation. We propose the Generalisation Index (GI), a metric that provides a clear, quantitative measure of a model's ability to transfer knowledge to new environments. Our framework achieves a near-perfect GI, providing strong empirical proof of its ability to learn domain-invariant representations, a critical step toward building truly deployable IoT sensing systems.}
\end{enumerate}

\section{Background and Related Work}

\subsection{Channel State Information (CSI) Fundamentals}
In modern wireless communication systems like IEEE 802.11n/ac, Orthogonal Frequency-Division Multiplexing (OFDM) is used to divide the frequency band into multiple orthogonal subcarriers, allowing for robust data transmission. Channel State Information (CSI) is a fine-grained measurement that describes the propagation characteristics of the signal from the transmitter to the receiver for each of these subcarriers~\cite{ma2019wifi}. The ability to extract this information from commercial WiFi devices, first demonstrated with specialised tools, has become the foundation for the entire field of WiFi sensing~\cite{wang2018device}.

To characterise the different paths, the multipath channel is modelled as a time-domain linear filter. Under time-invariant conditions, this is called the Channel Impulse Response (CIR), which can be expressed as:

\begin{equation}
    h(\tau) = \sum_{i=1}^{N} \alpha_i e^{-j\theta_i} \delta(t - \tau_i)
\end{equation}

Here, $N$ is the number of multipath components. For each path $i$, $\alpha_i$, $\theta_i$, and $\tau_i$ represent the amplitude, phase shift, and propagation delay, respectively, and $\delta(t)$ is the Dirac delta function. The Channel Frequency Response (CFR), which is what the CSI provides, is the Fourier transform of the CIR.

For a multiple input, multiple output (MIMO) OFDM system, the relationship between the transmitted signal vector $x$ and the received signal vector $y$ for a specific subcarrier $k$ can be modelled by the following equation:

\begin{equation}
    \mathbf{y}_k = \mathbf{H}_k \mathbf{x}_k + \mathbf{n}_k
\end{equation}

Here, $\mathbf{n}_k$ represents the Additive White Gaussian Noise (AWGN) vector. The core of the channel model is the matrix $\mathbf{H}_k$, which is the CFR. When the environment is dynamic due to human movement, the parameters of the CIR ($\alpha$, $\theta$, $\tau$) change over time. These changes are reflected in the CFR matrix $\mathbf{H}_k$, causing measurable fluctuations in the CSI. It is this direct sensitivity to environmental dynamics that allows CSI to be repurposed from a simple communication metric into a rich source of information for device-free sensing applications such as crowd-counting.

\subsection{Vision-Based Crowd-Counting}
Vision-based crowd-counting is the most established method for occupancy estimation, evolving from traditional techniques to the deep learning pipelines that are now standard. Early methods relied on regressing over hand-crafted features to map them to a crowd count~\cite{idrees2013multi}. However, these approaches were brittle and the field was revolutionised by the density map regression paradigm using convolutional neural networks (CNNs). A seminal work in this area was the Multi-Column CNN (MCNN)~\cite{zhang2016single}, which tackled the critical issue of scale variation by using parallel CNNs with different receptive fields. On the challenging ShanghaiTech Part A dataset, MCNN achieved a Mean Absolute Error (MAE) of 110.2, setting a new benchmark for deep learning methods. This was soon surpassed by architectures like CSRNet~\cite{li2018csrnet}, which used dilated convolutions to generate higher-resolution density maps. CSRNet achieved a landmark MAE of 68.2 on the same dataset, demonstrating a remarkable 38\% reduction in error and solidifying density map regression as the dominant methodology. Subsequent research focused on refining this approach by incorporating more sophisticated contextual reasoning, such as scale-adaptive modules in the Context-Aware Network (CAN)~\cite{liu2019context}. The latest state-of-the-art has continued to advance with models like CrowdFormer~\cite{yang2022crowdformer}, which leverages Vision Transformers to capture global context more effectively, further reducing the MAE to just 52.7 on the difficult JHU-Crowd++ benchmark.

Despite this impressive progress on in-domain benchmarks, the practical utility of these models is severely hampered by a fundamental failure to generalise. When a model trained on one dataset is tested on another without fine-tuning, its performance collapses. For example, a state-of-the-art model can see its MAE increase by over 200-300\% when transferred from the ShanghaiTech to the UCF-QNRF dataset, revealing that it has overfit to the specific camera angles, lighting, and backgrounds of the training domain~\cite{zhou2025crowd}. This brittleness is compounded by real-world operational challenges such as poor illumination and occlusions~\cite{kristoffersen2016pedestrian}. Most importantly, the reliance on continuous visual surveillance is often ethically contentious and legally problematic under data protection regulations like GDPR~\cite{mihaescu2024gdpr}, rendering cameras unsuitable for private spaces. These deeply rooted limitations have catalysed the search for alternative sensing modalities that are inherently robust and privacy-preserving.

\subsection{Crowd-Counting using WiFi Signals}
The use of WiFi Channel State Information (CSI) for device-free crowd-counting has been established as a viable, privacy-preserving alternative to vision-based methods. The fundamental principle is that human presence and movement alter RF signal propagation, and these changes, captured in the CSI, can be mapped to an occupant count~\cite{ma2019wifi,zhou2020device}. Initial research focused on proving this concept. For instance, Liu et al.~\cite{liu2017wicount} developed WiCount, a system using deep learning that achieved a mean accuracy of 92.8\% for counting up to 10 people in their test environment. To capture the rich spatio-temporal dynamics of CSI data, many early systems adopted hybrid deep learning models. A notable example is DeepCount~\cite{liu2019deepcount}, which combined CNNs and LSTMs and reported an average accuracy of 89.4\% for counting up to 5 people. The success of these initial systems spurred further research into more refined methodologies. For example, to better leverage the complementary nature of CSI data, Tian et al.~\cite{yan2025crowd} proposed a model using iterative attentional feature fusion to combine amplitude and phase information, achieving a counting accuracy of 95.2\% in their experimental setup. These studies solidified the potential of WiFi sensing, proving that under stable conditions, it could serve as a reliable method for fine-grained crowd analysis.

Despite these promising in-domain results, the critical challenge for WiFi sensing is its failure to generalise to new environments. The CSI features learnt by a model are highly dependent on the unique multipath "fingerprint" of a specific room, including its geometry and furnishings~\cite{zhu2025experience}. This leads to a severe performance collapse when a model is deployed in a new location. This problem is well-documented in the literature. A study by Li et al.~\cite{wu2022crowd} provides a stark example: a crowd-counting model that achieved 92.5\% accuracy when trained and tested in the same lab saw its performance plummet to just 37.5\% when transferred to a new conference room. This dramatic degradation highlights the primary obstacle to real-world adoption. Although domain adaptation techniques have been explored in related areas like gesture recognition~\cite{zhang2021widar3}, solving the cross-domain generalisation problem for crowd-counting remains the key unresolved challenge.

\subsection{Self-Supervised Domain Adaptation}
To overcome the critical challenge of domain shift, research has focused on domain adaptation techniques that learn domain-invariant representations. Early approaches sought to explicitly minimise the statistical distance between source and target domains, with methods like Deep Adaptation Networks (DAN)~\cite{long2015learning} using Maximum Mean Discrepancy to achieve, for example, an 8.2\% accuracy improvement over standard deep models on challenging cross-domain benchmarks. Another influential technique is Domain-Adversarial Training (DANN)~\cite{ganin2016domain}, which uses a domain classifier to force the feature extractor to learn domain-agnostic features, reducing the error rate on the MNIST-M benchmark from 44.1\% to 7.7\%.

More recently, self-supervised learning (SSL) has emerged as a more powerful and flexible paradigm. Contrastive frameworks like SimCLR~\cite{chen2020simple} learn robust representations from unlabelled data by maximising the agreement between augmented views of the same sample. This approach proved highly effective, achieving 76.5\% top-1 accuracy on ImageNet with a linear classifier, rivaling supervised pre-training. Non-contrastive methods like Bootstrap Your Own Latent (BYOL)~\cite{grill2020bootstrap} further refined this, achieving 79.6\% top-1 accuracy on the same benchmark without requiring negative samples.

The application of SSL to WiFi sensing is now recognised as a key research direction for building robust systems~\cite{radwan2025tutorial}. By applying these techniques to CSI data, models can learn to focus on the intrinsic patterns of human activity while ignoring domain-specific environmental noise. The results have been compelling. For instance, Yan et al.~\cite{yan2025wi} developed Wi-SFDAGR, a source-free domain adaptation system for gesture recognition. On a cross-domain task, their method improved accuracy from a baseline of 47.1\% to 89.4\%—a dramatic recovery of performance. Similarly, Zhang et al.~\cite{zhang2018crosssense} proposed CrossSense, a system that disentangles domain-specific and domain-invariant features, improving the average cross-site activity recognition accuracy from 53.3\% to 83.1\%. These results demonstrate that self-supervised and domain adaptation methods are the state-of-the-art solution for learning the domain-invariant features required for robust, real-world crowd-counting systems.

\section{Proposed Framework}

\begin{figure*}[t]
    \centering
    \makebox[\textwidth][c]{%
        \includegraphics[width=0.9\textwidth]{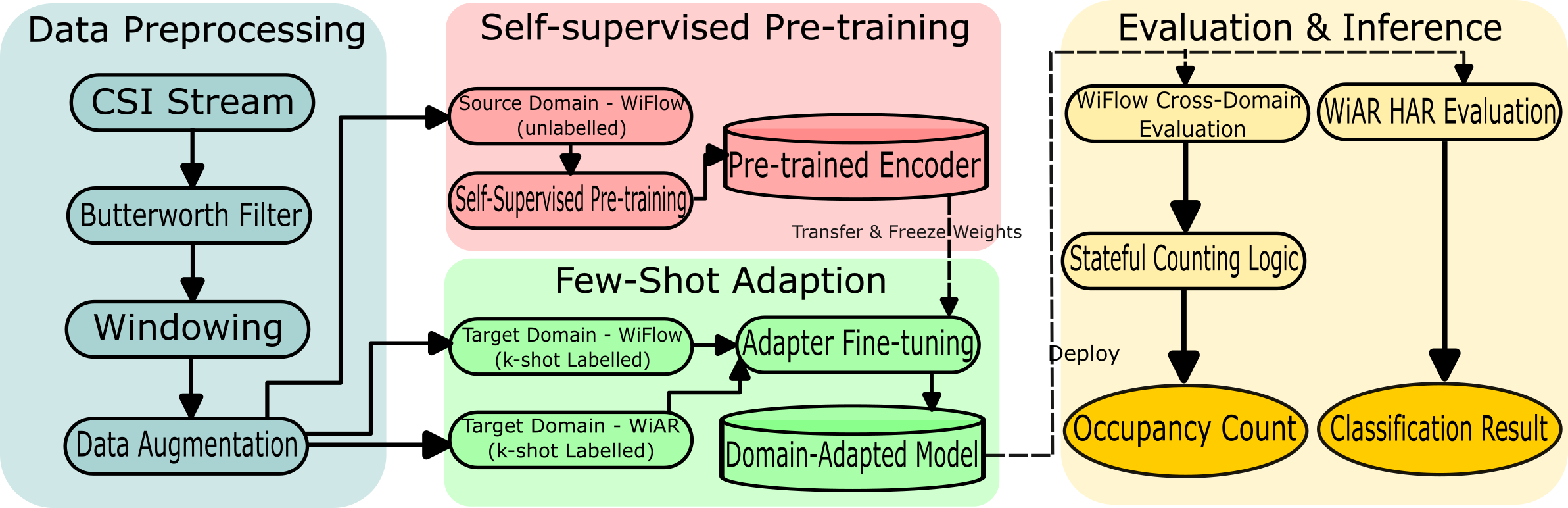}%
    }
    \captionsetup{justification=centering} 
    \caption{Overview of the proposed system pipeline.}
    \label{fig:wide}
\end{figure*}

\subsection{Overview}
Our proposed framework, illustrated in Fig~\ref{fig:wide}, presents a comprehensive pipeline for cross-domain WiFi-based crowd counting that begins with raw CSI data and concludes with robust occupancy estimation. The initial Data Preprocessing stage takes the raw CSI stream and applies a Butterworth filter to remove high-frequency noise, followed by segmenting the signal into fixed-size windows. To prepare for the subsequent training stages and enhance model robustness, data augmentation techniques—specifically jitter, scaling, and masking—are applied to these windows.

The core of our methodology consists of two training phases. First, in the Self-Supervised Pre-training stage, the processed, unlabelled data from a source domain is used to train our encoder model via a contrastive learning objective. This step produces a powerful, pre-trained encoder that captures general, domain-agnostic features of human motion. Next, in the Few-Shot Adaptation stage, the learnt weights from the encoder are transferred and frozen. The model is then efficiently adapted to a new target domain by fine-tuning only lightweight 
\textit{Adapter} modules using a small, k-shot labelled dataset from the target environment. This results in a domain-adapted model ready for deployment.

Finally, the Evaluation and Inference stage demonstrates the model's capabilities on two distinct downstream tasks. The deployed model processes a test data stream to generate a sequence of event predictions. For cross-domain evaluation on the WiAR dataset, these predictions serve as the final output for Human Activity Recognition (HAR) classification. For our primary task on our WiFlow dataset, the model first classifies events as `enter', `exit', or `no event'. These classifications are then fed into a stateful counting logic module, which translates the event sequence into a final, stable occupancy count by filtering spurious detections.

\subsection{Data Preprocessing and Augmentation}\label{sec:preprocessing}
Raw CSI data is susceptible to environmental noise and requires significant structure before it can be used for effective feature learning. To address this, we developed a multi-stage data processing pipeline designed to denoise the signal, segment it into meaningful temporal windows, and generate augmented views for contrastive pre-training.

The initial preprocessing step mitigates the influence of high-frequency noise. Given that human body movements typically occur at low frequencies (generally below 6 Hz)~\cite{wang2017device}, we apply a 4th-order Butterworth low-pass filter to the raw CSI amplitude streams. The filter is designed with a cutoff frequency $f_c$ of 8 Hz to preserve the activity-related signal while attenuating noise. For our digital implementation with a sampling rate $F_s$ of 100 Hz, the normalized cutoff frequency $W_n$ required for the filter design is calculated as follows: 

\begin{equation} 
W_n = \frac{f_c}{F_s / 2} = \frac{8}{100 / 2} = 0.16 
\end{equation} 

\begin{figure}[!htbp]
\centering
\includegraphics[width=\columnwidth]{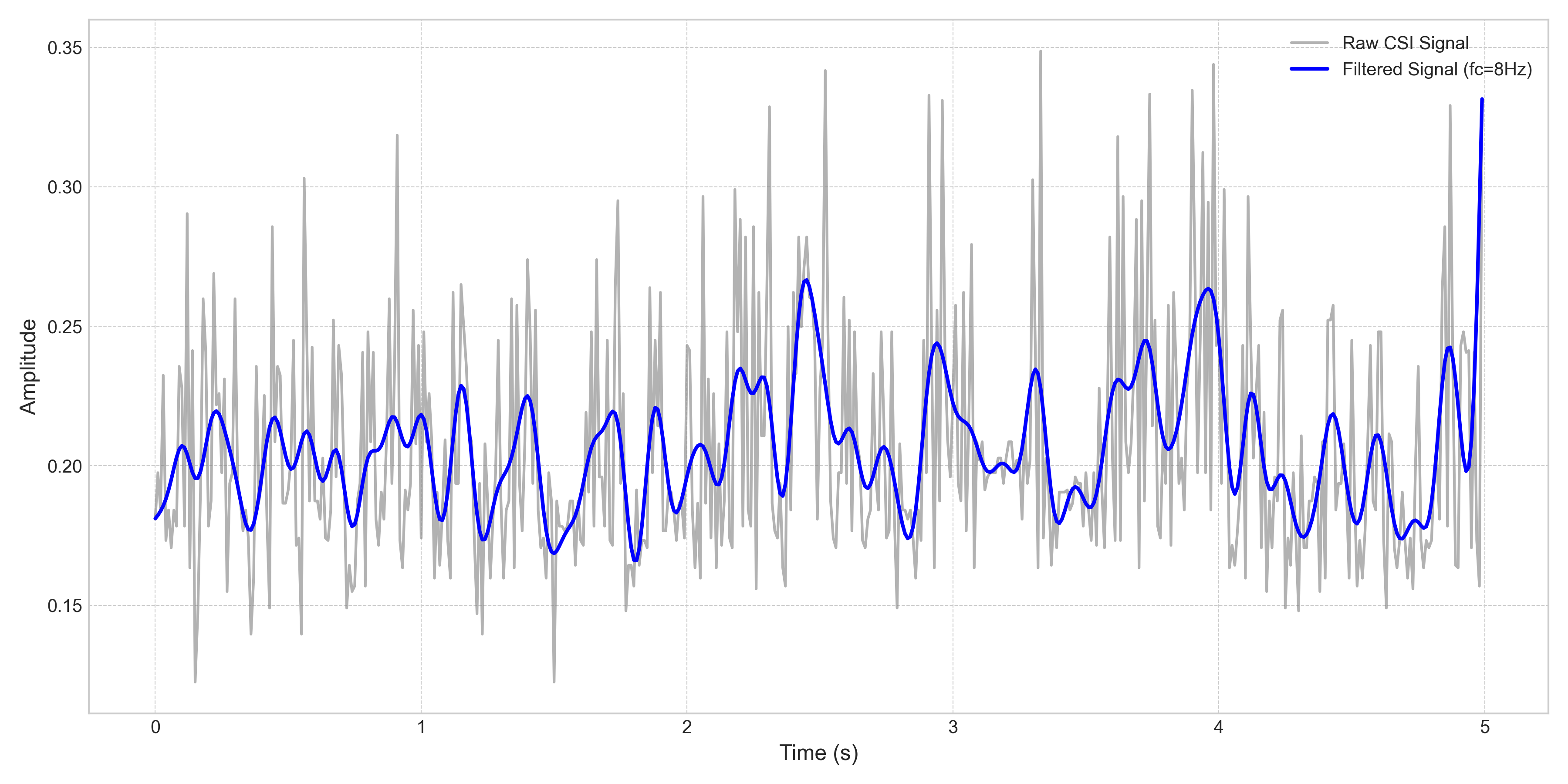}
\caption{A raw CSI waveform shown before and after denoising with a 4th-order Butterworth filter.}
\label{fig2}
\end{figure}

This normalised frequency is used to define the filter's transfer function. The filtering operation can be represented as $X_{\text{filtered}} = \text{LPF}(X_{\text{raw}})$, where $X_{\text{raw}}$ is the raw CSI time series and $\text{LPF}(\cdot)$ denotes the low-pass filtering function. This process effectively smooths the signal, enhancing the underlying patterns corresponding to human activity, as illustrated in Fig~\ref{fig2}.

Following denoising, the continuous signal is segmented into fixed-length windows. A sliding window of length $W=100$ (1 second) is applied with a step size of $S=50$, resulting in a 50\% overlap between consecutive windows. This windowing process transforms the time series $X_{\text{filtered}}$ into a set of instances $\{w_i\}$, where each window $w_i$ corresponds to the segment $X_{\text{filtered}}[iS : iS+W]$. For the supervised fine-tuning stage, we enforce label purity by retaining only those windows that contain a single, consistent event type (`enter', `exit', or `no\_event'), ensuring the model is trained on unambiguous examples.

The cornerstone of our self-supervised learning approach is the generation of semantically invariant views from each window $w$ for contrastive learning. This is achieved through a policy of strong data augmentation, a technique proven to be effective for learning robust time-series representations~\cite{yue2022ts2vec}. For each window $w$, we generate a positive pair $(w', w'')$ by applying a composition of transformations. To improve robustness to sensor noise, we introduce additive Gaussian noise $\epsilon$, where $\epsilon \sim \mathcal{N}(0, \sigma_j^2)$, such that an augmented view $w'(t)$ is given by: \begin{equation} 
w'(t) = w(t) + \epsilon 
\end{equation} 

To ensure invariance to signal magnitude, we apply a random scaling factor $\alpha$, where $\alpha \sim \mathcal{N}(1, \sigma_s^2)$, according to the transformation: 
\begin{equation} 
w'(t) = \alpha \cdot w(t) 
\end{equation} 

Finally, to encourage the learning of holistic features over local temporal patterns, we apply a permutation operator $P(\cdot)$, which partitions a window $w$ into $k$ segments, where $k$ is randomly chosen from $\{2, 3, 4, 5\}$, and reassembles them in a random order.

By training the model to recognise $w'$ and $w''$ as originating from the same source window, the encoder is compelled to learn a feature representation that is invariant to these transformations. This pre-training process is critical for developing a powerful and generalizable model that can be effectively adapted to new tasks with limited labelled data.

\subsection{Model Architecture}

\begin{table}[t]
\centering
\caption{The detailed configuration of the CSI-ResNet-A model. The full architecture of the first residual block in each of the three layers is shown. Subsequent blocks within the same layer repeat the same structure. In the 'Details' column, 'k' denotes kernel size, 's' denotes stride, and 'p' denotes padding.}
\label{tab:table1}
\footnotesize
\setlength{\tabcolsep}{3pt}
\begin{tabular*}{\columnwidth}{@{\extracolsep{\fill}} l l r l}
\toprule
\textbf{Layer} & \textbf{Output Dim.} & \textbf{Params \#} & \textbf{Details} \\ 
\midrule
Input & 52 x 100 & 0 & \\ 
\midrule
\multicolumn{4}{@{}l}{\textit{Initial Convolutional Block}} \\
\hspace{1em} - conv1 (Conv1D) & 64 x 50 & 23,360 & k=7, s=2, p=3 \\
\hspace{1em} - bn1 (BatchNorm1d) & 64 x 50 & 128 & \\
\hspace{1em} - relu (ReLU) & 64 x 50 & 0 & \\
\hspace{1em} - pool1 (MaxPool1d) & 64 x 25 & 0 & k=3, s=2, p=1 \\ 
\midrule
\multicolumn{4}{@{}l}{\textbf{Residual Block 1.1}} \\
\hspace{1em} - conv1 & 64 x 25 & 12,352 & k=3, s=1, p=1 \\
\hspace{1em} - bn1 & 64 x 25 & 128 & \\
\hspace{1em} - relu1 & 64 x 25 & 0 & \\
\hspace{1em} - conv2 & 64 x 25 & 12,352 & k=3, s=1, p=1 \\
\hspace{1em} - bn2 & 64 x 25 & 128 & \\
\hspace{1em} - Squeeze-and-Excitation & 64 x 25 & 512 & reduction=16 \\
\hspace{1em} - adapter & 64 x 25 & 2,128 & bottleneck=16 \\
\hspace{1em} - add\_shortcut & 64 x 25 & 0 & \\
\hspace{1em} - relu2 & 64 x 25 & 0 & \\
\textbf{Residual Block 1.2} & 64 x 25 & 27,600 & \textit{Repeats Block 1.1} \\
\midrule
\multicolumn{4}{@{}l}{\textbf{Residual Block 2.1}} \\
\hspace{1em} - conv1 & 128 x 13 & 24,704 & k=3, s=2, p=1 \\
\hspace{1em} - bn1 & 128 x 13 & 256 & \\
\hspace{1em} - relu1 & 128 x 13 & 0 & \\
\hspace{1em} - conv2 & 128 x 13 & 49,280 & k=3, s=1, p=1 \\
\hspace{1em} - bn2 & 128 x 13 & 256 & \\
\hspace{1em} - Squeeze-and-Excitation & 128 x 13 & 2,048 & reduction=16 \\
\hspace{1em} - adapter & 128 x 13 & 4,240 & bottleneck=16 \\
\hspace{1em} - add\_shortcut & 128 x 13 & 0 & \\
\hspace{1em} - relu2 & 128 x 13 & 0 & \\
\textbf{Residual Block 2.2} & 128 x 13 & 86,448 & \textit{Repeats Block 2.1} \\
\midrule
\multicolumn{4}{@{}l}{\textbf{Residual Block 3.1}} \\
\hspace{1em} - conv1 & 256 x 7 & 98,560 & k=3, s=2, p=1 \\
\hspace{1em} - bn1 & 256 x 7 & 512 & \\
\hspace{1em} - relu1 & 256 x 7 & 0 & \\
\hspace{1em} - conv2 & 256 x 7 & 196,864 & k=3, s=1, p=1 \\
\hspace{1em} - bn2 & 256 x 7 & 512 & \\
\hspace{1em} - Squeeze-and-Excitation & 256 x 7 & 8,192 & reduction=16 \\
\hspace{1em} - adapter & 256 x 7 & 8,464 & bottleneck=16 \\
\hspace{1em} - add\_shortcut & 256 x 7 & 0 & \\
\hspace{1em} - relu2 & 256 x 7 & 0 & \\
\textbf{Residual Block 3.2} & 256 x 7 & 346,192 & \textit{Repeats Block 3.1} \\
\midrule
\multicolumn{4}{@{}l}{\textit{Output Head}} \\
\hspace{1em} Global Avg Pooling & 256 x 1 & 0 & \\
\hspace{1em} Squeeze & 256 & 0 & \\
\hspace{1em} Fully Connected & 128 & 32,896 & in=256, out=128 \\ 
\midrule
\multicolumn{2}{@{}l}{\textbf{Total Parameters}} & \multicolumn{2}{r@{}}{\textbf{1,092,806}} \\
\multicolumn{2}{@{}l}{\textbf{Trainable Parameters}} & \multicolumn{2}{r@{}}{\textbf{1,092,806}} \\
\bottomrule
\end{tabular*}
\end{table}

\begin{figure*}[t]
    \centering
    \makebox[\textwidth][c]{%
        \includegraphics[width=0.9\textwidth]{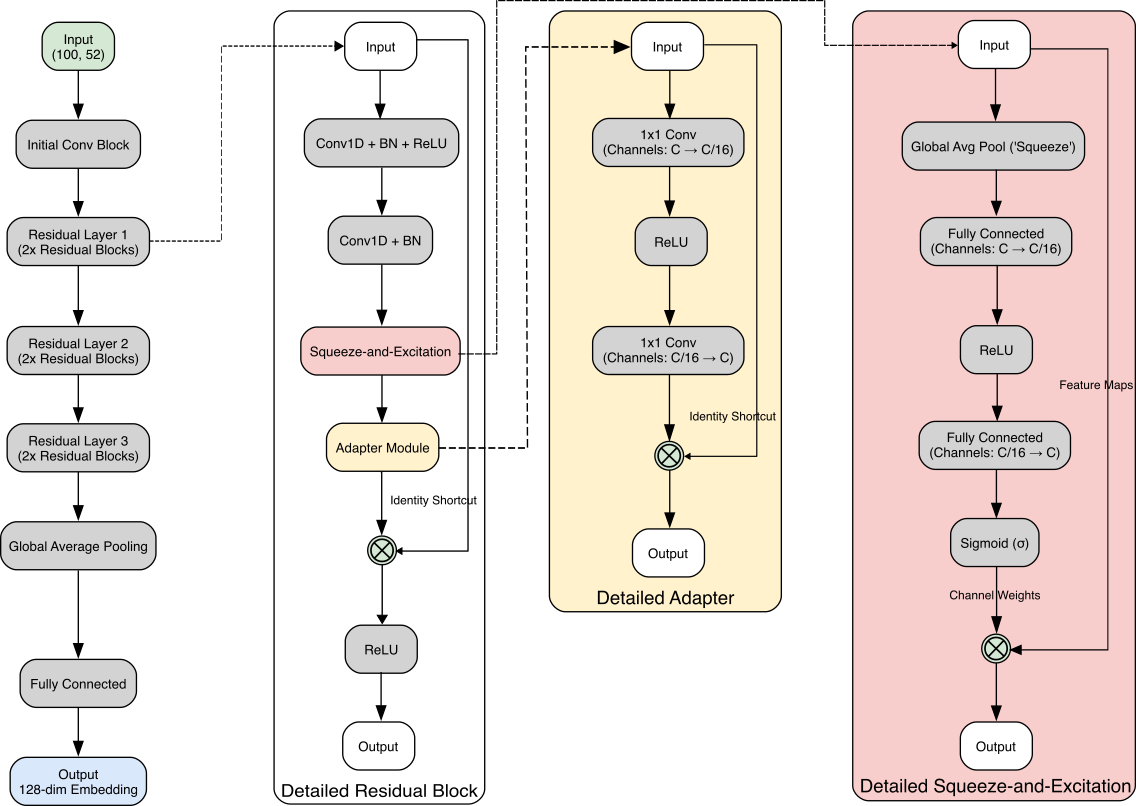}%
    }
    \captionsetup{justification=centering} 
    \caption{The proposed CSI-ResNet-A architecture.}
    \label{fig:fig3}
\end{figure*}

This section provides a detailed mathematical and structural description of our proposed neural network, CSI-ResNet-A. The architecture is designed to learn robust feature representations from CSI data while remaining highly efficient for adaptation to new tasks. The model is built upon a ResNet backbone, enhanced with channel-wise attention and lightweight adapter modules. A comprehensive overview is presented in Fig~\ref{fig:fig3}, with layer configurations detailed in Table~\ref{tab:table1}.

\subsubsection{CSI Encoder (ResNet Backbone)}
The core of our model is a deep 1D Convolutional Neural Network (CNN) based on the ResNet architecture~\cite{he2016deep}, which has proven effective for time-series classification~\cite{ismail2019deep}. Its residual connections are critical for training deep networks by mitigating the vanishing gradient problem.

Let $\mathbf{X} \in \mathbb{R}^{B \times L \times C}$ be an input batch of $B$ windows, where $L=100$ is the window length and $C=52$ is the number of CSI subcarriers. The input is first permuted to $\mathbf{X}_{\text{in}} \in \mathbb{R}^{B \times C \times L}$ and passed through an initial convolutional block. A 1D convolution is a feature detector that applies a kernel $\mathbf{K} \in \mathbb{R}^{C_{\text{out}} \times C_{\text{in}} \times k}$ across the temporal dimension of an input $\mathbf{x} \in \mathbb{R}^{C_{\text{in}} \times L}$: 

\begin{equation} \text{Conv1D}(\mathbf{x})j = \mathbf{b}j + \sum{i=1}^{C{\text{in}}} (\mathbf{K}_{j,i} * \mathbf{x}_i) \end{equation} 

where $*$ denotes the valid cross-correlation, $\mathbf{K}_{j,i}$ is the kernel for the $j$-th output channel and $i$-th input channel, and $\mathbf{b}_j$ is a bias term. The initial block creates a feature map $\mathbf{X}_0$: 

\begin{equation} \mathbf{X}_0 = \text{MaxPool1D}(\delta(\text{BN}(\text{Conv1D}(\mathbf{X}_{\text{in}})))) \end{equation} 

where $\delta$ is the ReLU activation function, $\text{BN}$ is Batch Normalization~\cite{ioffe2015batch}, and MaxPool1D downsamples the feature map.

The encoder body consists of a series of residual blocks. The output of the $\ell$-th block, $\mathbf{x}_{\ell+1}$, is defined by its input $\mathbf{x}_{\ell}$ and a residual function $\mathcal{F}$: 

\begin{equation} \mathbf{x}{\ell+1} = \delta(\mathcal{F}(\mathbf{x}{\ell}, {\mathbf{W}i}{\ell}) + \mathcal{W}s\mathbf{x}{\ell}) \end{equation} 

where $\{\mathbf{W}_i\}_{\ell}$ are the weights of the $\ell$-th block and $\mathcal{W}_s$ is a linear projection to match dimensions. The residual function $\mathcal{F}$ is a sequence of two 1D convolutional blocks. The final feature representation $\mathbf{e} \in \mathbb{R}^{B \times D_e}$ (where $D_e=128$) is produced by applying a global average pooling (GAP) layer, followed by a fully connected (FC) layer: 

\begin{equation} \mathbf{e} = \text{FC}(\text{GAP}(\mathbf{x}_{\text{final}})) \end{equation}

\subsubsection{Squeeze-and-Excitation (SE) Blocks}
To enhance representational power, we integrate Squeeze-and-Excitation (SE) modules within each residual block~\cite{hu2018squeeze}. SE blocks introduce channel-wise attention, allowing the model to dynamically re-weigh feature maps.

First, the Squeeze operation aggregates a feature map $\mathbf{U} \in \mathbb{R}^{C' \times L'}$ into a channel descriptor vector $\mathbf{z} \in \mathbb{R}^{C'}$ using global average pooling: 

\begin{equation} z_c = \frac{1}{L'} \sum_{i=1}^{L'} u_c(i) \end{equation} 

where $u_c(i)$ is the value at temporal position $i$ of channel $c$. Second, the Excitation operation generates channel-wise weights $\mathbf{s} \in \mathbb{R}^{C'}$ using a two-layer bottleneck MLP: 

\begin{equation} \mathbf{s} = \sigma(\mathbf{W}_2 \delta(\mathbf{W}_1 \mathbf{z})) \end{equation} 

where $\mathbf{W}_1 \in \mathbb{R}^{\frac{C'}{r} \times C'}$ and $\mathbf{W}_2 \in \mathbb{R}^{C' \times \frac{C'}{r}}$ are the MLP weights, $\delta$ is ReLU, $\sigma$ is the sigmoid function, and $r=16$ is the reduction ratio.

Finally, the output of the SE module, $\tilde{\mathbf{X}}$, is obtained by rescaling $\mathbf{U}$ with the weights $\mathbf{s}$: 

\begin{equation} \tilde{\mathbf{X}}_c = s_c \cdot \mathbf{U}_c \end{equation} 

where $\mathbf{U}_c$ is the $c$-th channel of $\mathbf{U}$ and $s_c$ is the $c$-th element of $\mathbf{s}$.

\subsubsection{Adapter Modules}
To facilitate parameter-efficient fine-tuning, we incorporate lightweight Adapter modules~\cite{houlsby2019parameter}. During adaptation, only the Adapter weights are trained. As shown in Fig. 3, our Adapter is inserted after the SE block in each residual block. It processes an input feature map $\mathbf{x}_{\text{in}} \in \mathbb{R}^{C' \times L'}$ with a residual bottleneck architecture: 

\begin{equation} \mathbf{x}_{\text{out}} = (\mathbf{K}_{\text{up}} * (\delta(\mathbf{K}_{\text{down}} * \mathbf{x}_{\text{in}} + \mathbf{b}_{\text{down}})) + \mathbf{b}_{\text{up}}) + \mathbf{x}_{\text{in}} \end{equation} 

where $\mathbf{K}_{\text{down}}$ and $\mathbf{K}_{\text{up}}$ are the kernels of 1x1 convolutions that project the C' channels down to a bottleneck size ($C'/16$) and back up, respectively. $\mathbf{b}_{\text{down}}$ and $\mathbf{b}_{\text{up}}$ are the corresponding bias vectors. The kernels are initialized such that the adapter initially performs an identity mapping.

\subsubsection{Classification Head}
For the final event classification task, a simple linear \textit{ClassificationHead} is placed on top of the pre-trained and adapted \textit{CSIEncoder}. This head consists of a single fully connected layer that maps the 128-dimensional feature embedding $\mathbf{e}$ to a 3-dimensional logit vector $\mathbf{y}_{\text{logits}}$, corresponding to the three event classes: `enter', `exit', and `no\_event'. The operation is given by: 

\begin{equation}
\mathbf{y}_{\text{logits}} = \mathbf{W}_{\text{head}}\mathbf{e} + \mathbf{b}_{\text{head}}
\end{equation}

where $\mathbf{W}_{\text{head}} \in \mathbb{R}^{3 \times 128}$ and $\mathbf{b}_{\text{head}} \in \mathbb{R}^{3}$ are the learnable weights and biases. The final class probabilities $\mathbf{p} \in \mathbb{R}^3$ are obtained via the softmax function: 

\begin{equation} p_i = \frac{\exp(y_{\text{logits}, i})}{\sum_{j=1}^{3} \exp(y_{\text{logits}, j})} \end{equation} 

where $p_i$ is the predicted probability for class $i$.

\subsection{Training and Adaptation Pipeline}
This section describes the two-stage pipeline used to train and adapt our model. The process begins with a self-supervised pre-training phase on a large, unlabelled source dataset to learn robust features, followed by a highly efficient fine-tuning phase to adapt the model to a new target domain with minimal labelled data.

\subsubsection{Self-Supervised Pre-training}
The primary goal of the pre-training stage is to learn a powerful and generalisable feature representation from unlabelled CSI data. To achieve this, we employ a contrastive learning framework inspired by SimCLR~\cite{chen2020simple}. This approach trains the \textit{CSIEncoder} to produce similar embeddings for different augmented "views" of the same input window, while pushing apart the embeddings of different windows.

For each input window $\mathbf{x}$ in a batch, we generate a positive pair of augmented views, $(\mathbf{x}_i, \mathbf{x}_j)$, using the strong augmentation policy detailed in Section~\ref{sec:preprocessing}. These views are then passed through the \textit{CSIEncoder} ($f(\cdot)$) and a non-linear \textit{ProjectionHead} ($g(\cdot)$) to obtain projected latent representations, $\mathbf{z}_i = g(f(\mathbf{x}_i))$ and $\mathbf{z}_j = g(f(\mathbf{x}_j))$. The projection head is only used during this pre-training phase and is discarded afterward.

The model is trained by minimising the Normalized Temperature-scaled Cross-Entropy (NT-Xent) loss~\cite{oord2018representation}. For a positive pair $(\mathbf{z}_i, \mathbf{z}_j)$, the loss is defined as: 

\begin{equation} \mathcal{L}_{i,j} = -\log \frac{\exp(\text{sim}(\mathbf{z}i, \mathbf{z}j) / \tau)}{\sum{k=1}^{2N} \mathbb{1}{[k \neq i]} \exp(\text{sim}(\mathbf{z}_i, \mathbf{z}_k) / \tau)} \end{equation} 

where $\text{sim}(\mathbf{u}, \mathbf{v}) = \mathbf{u}^\top \mathbf{v} / \|\mathbf{u}\| \|\mathbf{v}\|$ is the cosine similarity, $\tau$ is a temperature parameter that scales the logits, $\mathbb{1}_{[k \neq i]}$ is an indicator function evaluating to 1 if $k \neq i$, and $2N$ is the total number of augmented samples in the batch. The total loss is computed over all positive pairs in the batch.

\begin{figure}[!t]
\centering
\includegraphics[width=\columnwidth]{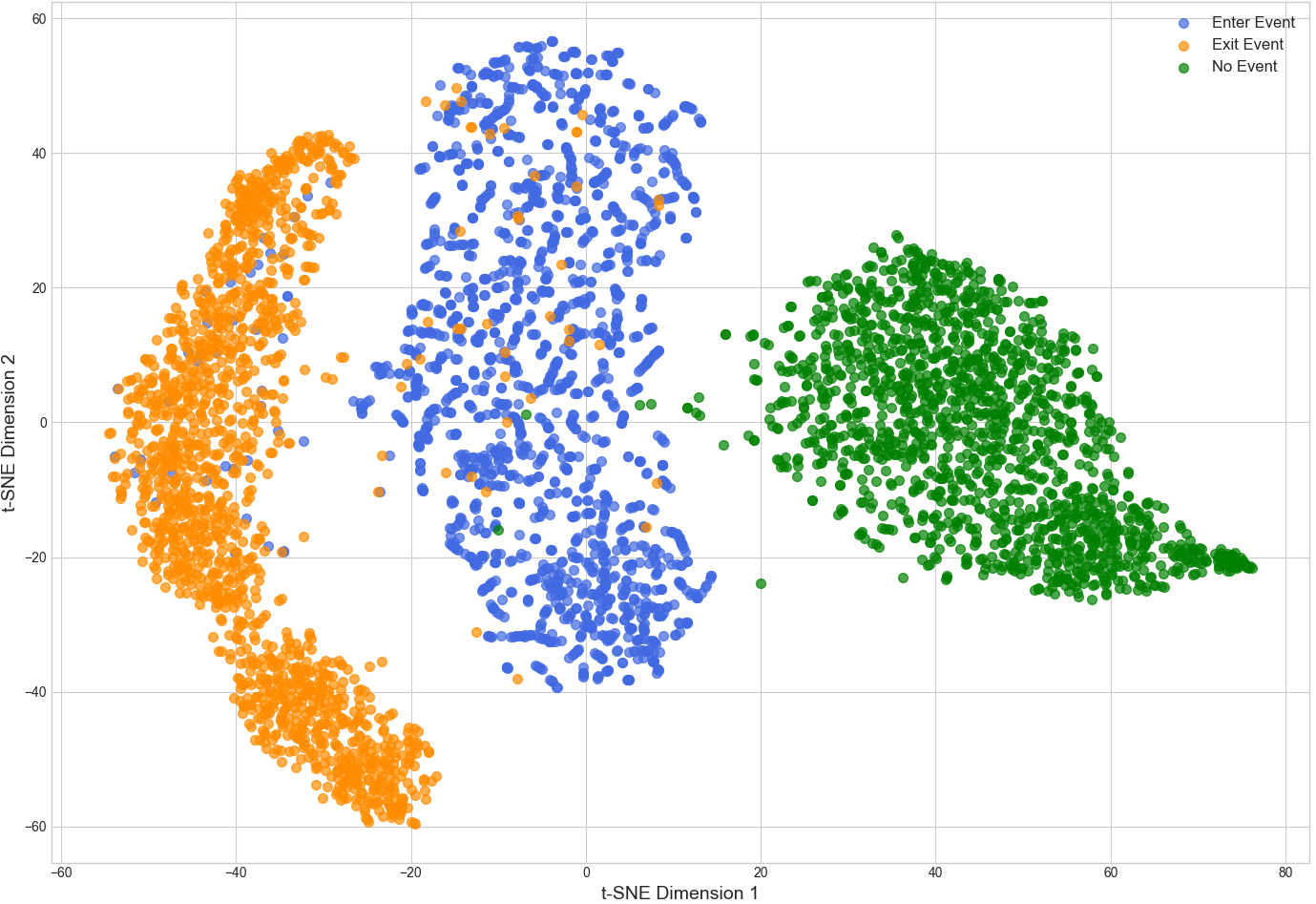}
\caption{t-SNE visualization of the pre-trained CSI embeddings.}
\label{fig4}
\end{figure}

This objective effectively trains the encoder to be invariant to the applied augmentations, forcing it to learn the essential underlying structure of the CSI signals. The quality of the learnt feature space is visualised in Fig~\ref{fig4}, where a t-SNE projection of the embeddings from a test set reveals distinct clusters for the `enter', `exit', and `no\_event' classes, even though the model was trained without any labels.

\subsubsection{Few-Shot Cross-Domain Fine-Tuning}
To adapt the model to a new target environment or dataset (e.g., from our WiFlow data to the public WiAR dataset), we employ a two-stage adaptation strategy designed for maximum data and parameter efficiency.

First, for datasets with a significant domain shift, we perform unsupervised domain adaptation using an Adversarial Discriminative Domain Adaptation (ADDA) approach~\cite{tzeng2017adversarial}. This step aligns the feature distributions of the source ($S$) and target ($T$) domains without using any target labels. A target encoder, $G_T$, is trained to produce embeddings that can fool a domain discriminator, $D$, which is simultaneously trained to distinguish between source embeddings from $G_S$ and target embeddings from $G_T$. The discriminator's loss is a standard binary cross-entropy: 
\begin{equation}
\begin{split}
    \min_{D} \mathcal{L}_{\text{adv}_D} ={}& - \mathbb{E}_{\mathbf{x}_S \sim S}[\log D(G_S(\mathbf{x}_S))] \\
    & - \mathbb{E}_{\mathbf{x}_T \sim T}[\log(1 - D(G_T(\mathbf{x}_T)))]
\end{split}
\end{equation}

The target encoder is trained to maximise this discriminator loss, effectively minimising: 

\begin{equation} \min{G_T} \mathcal{L}{\text{adv}{G_T}} = - \mathbb{E}{\mathbf{x}_T \sim T}[\log D(G_T(\mathbf{x}_T))] \end{equation} 

This adversarial game encourages the target encoder to produce embeddings that are indistinguishable from the source embeddings, thus aligning the feature spaces.

Second, we perform parameter-efficient fine-tuning on the adapted encoder using a very small number of labelled examples from the target domain ($k$-shot learning). To prevent catastrophic forgetting and minimise the number of trainable parameters, we freeze the entire backbone of the \textit{CSIEncoder}. Only the lightweight \textit{Adapter} modules and a newly initialised \textit{ClassificationHead} are trained. The model is fine-tuned by minimising the standard cross-entropy loss $\mathcal{L}_{\text{CE}}$ on the $k$ labelled examples: 

\begin{equation} \mathcal{L}_{\text{CE}} = - \sum{i=1}^{N} \mathbf{y}_i \log(\hat{\mathbf{y}}_i) \end{equation} 

where $\mathbf{y}_i$ is the one-hot encoded true label and $\hat{\mathbf{y}}_i$ is the predicted probability distribution from the softmax output of the \textit{ClassificationHead}. This two-step process allows for rapid and robust adaptation of the model to new environments with minimal data and computational overhead.

\subsection{Stateful Occupancy Counting}
The final stage of our framework translates the sequence of classified events (`enter', `exit', `no\_event') from the model into a stable, real-time occupancy count. Raw model outputs can be noisy, with spurious classifications that would lead to an erratic and inaccurate count if processed directly. To address this, we implement a robust state machine that ensures only persistent, deliberate events are registered.

\subsubsection{Debouncing and Cooldown Logic}
\begin{figure}[b]
\centering
\includegraphics[width=\columnwidth]{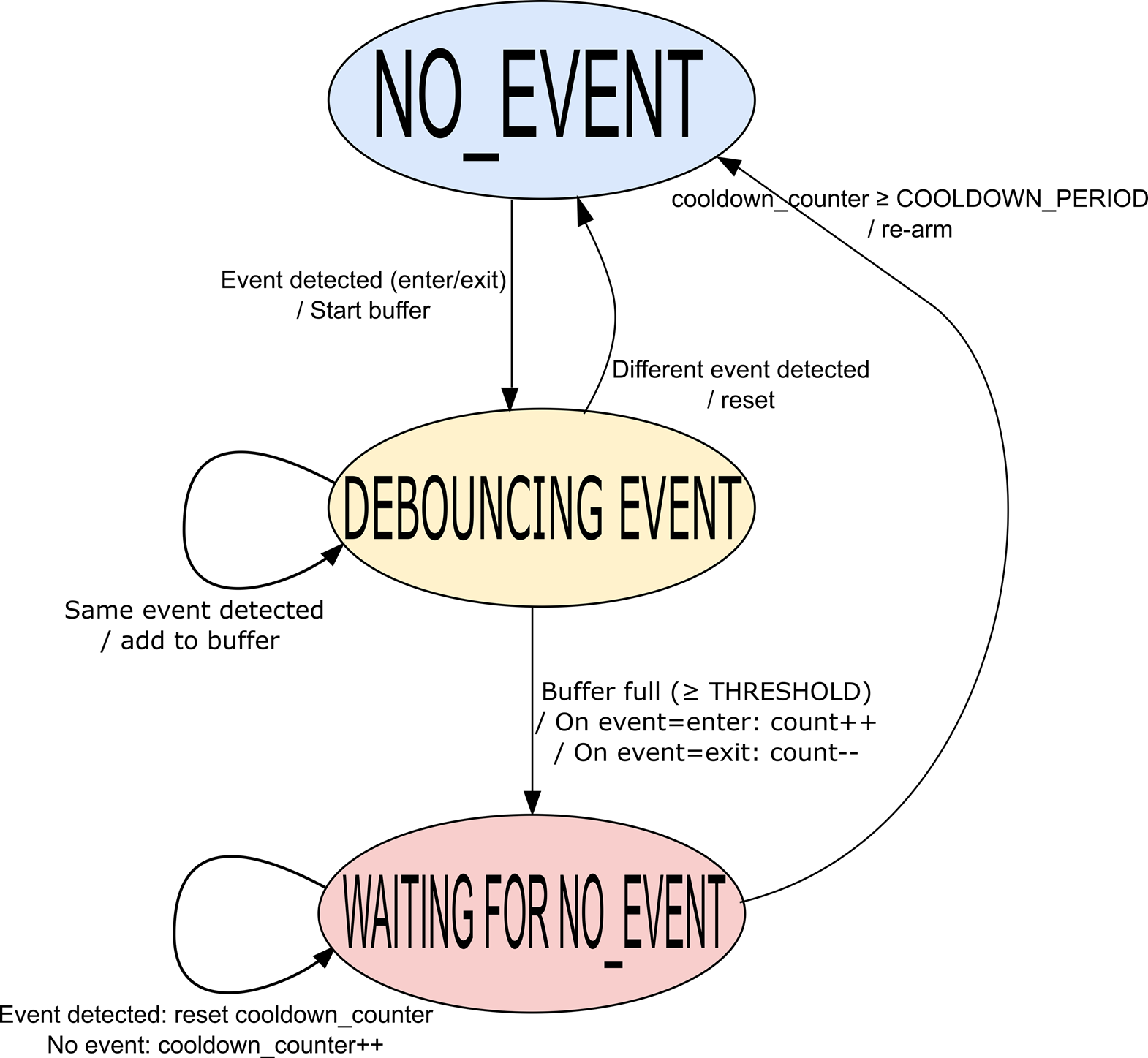}
\caption{State machine for robust occupancy counting.}
\label{fig5}
\end{figure}

The core of our counting mechanism is a finite state machine, illustrated in Fig~\ref{fig5}, which governs how the occupancy count is updated. This state machine is designed to be resilient to transient noise and prevent double-counting of single events. It operates using three primary states: NO\_EVENT, DEBOUNCING EVENT, and WAITING FOR NO\_EVENT.

Initially, the system is in the no\_event state. When a non-zero event (either `enter' or `exit') is first detected, the system transitions to the DEBOUNCING EVENT state. In this state, it accumulates consecutive, identical event classifications in a buffer. An event is only confirmed and the occupancy count updated if the number of identical, sequential classifications meets a predefined EVENT\_THRESHOLD. This debouncing mechanism effectively filters out fleeting, noisy predictions that do not represent a sustained activity. If a different event type is detected before the threshold is met, the buffer is reset, and the system returns to the no\_event state.

Once an event is confirmed and the count is updated, the system transitions to the WAITING FOR NO\_EVENT state. In this state, it will not register any new events until a COOLDOWN\_PERIOD of consecutive `no\_event' classifications has been observed. This cooldown logic is crucial for preventing a single, prolonged activity (like a person walking through a doorway) from being counted multiple times. After the cooldown period is complete, the system is re-armed by returning to the no\_event state, ready to detect the next valid event. This stateful, logic-based approach ensures a robust and reliable final occupancy estimate.

\section{Experiments}

\subsection{Implementation Details}
\subsubsection{Framework and Hardware} The sensing hardware comprised two ESP32-WROOM-32U microcontrollers with external antennas, flashed using the CSI tool from Hernandez and Bulut~\cite{Hern2006:Lightweight}. These were connected to a Raspberry Pi for data logging. All subsequent data processing and model training were performed on a 13-inch MacBook Pro (2019) with a 1.4 GHz Quad-Core Intel Core i5 processor and 8 GB of RAM, using the PyTorch framework on the CPU.

\subsubsection{Preprocessing and Augmentation Parameters} Raw CSI amplitude data was filtered using a 4th-order Butterworth low-pass filter with an 8 Hz cutoff frequency. The signal was then segmented into windows of 100 timesteps with a 50-timestep step size (50\% overlap). For contrastive learning, a series of augmentations were applied, including additive Gaussian noise with a standard deviation of 0.03 (Jitter), multiplicative scaling with a factor drawn from a normal distribution with a mean of 1.0 and a standard deviation of 0.1 (Scale), and the random shuffling of up to 5 segments of the window (Permutation).

\subsubsection{Pre-training Hyperparameters} The \textit{CSIEncoder} was pre-trained on our WiFlow dataset for 50 epochs using the Adam optimiser with a learning rate of $1 \times 10^{-3}$ and a batch size of 128. The NT-Xent contrastive loss function was configured with a temperature $\tau$ of 0.1.

\subsubsection{Evaluation Hyperparameters} For the WiFlow dataset, we evaluated four distinct methods to provide a comprehensive comparison. The Traditional Baseline involved extracting statistical features and training a Random Forest classifier. The Supervised Baseline trained the \textit{CSIEncoder} and classification head from scratch. The Linear Probe method involved training a simple Logistic Regression classifier on top of the frozen features from our pre-trained encoder. Finally, our proposed Fine-Tuning method updated the final block of the pre-trained encoder and the classification head. For the fine-tuning and supervised models, we trained for 25 epochs with a batch size of 16, using the Adam optimizer with discriminative learning rates of $1 \times 10^{-4}$ for the encoder layers and $1 \times 10^{-3}$ for the classification head. All k-shot evaluations were repeated 10 times.

For the WiAR dataset, we evaluated two main transfer learning scenarios. The Source-Only scenario used the encoder pre-trained on WiFlow without any adaptation. The ADDA scenario first adapted the encoder to the WiAR domain for 50 epochs. Both scenarios were then evaluated using a linear\_probe and a fine\_tune approach. The fine-tuning was performed for 75 epochs with a batch size of 32, using the Adam optimiser and a Focal Loss criterion to handle class imbalance.

\subsubsection{State Machine Parameters} The stateful counter was configured with an EVENT\_THRESHOLD of 5 consecutive windows to confirm an event and a COOLDOWN\_PERIOD of 10 consecutive windows to re-arm the system after an event.

\subsection{Datasets}
To rigorously evaluate the performance and Generalisation capabilities of our framework, we utilise two distinct datasets: our custom-collected WiFlow dataset for the primary task of crowd counting, and the public WiAR dataset to test cross-domain transferability to a related, but different, activity recognition task.

The WiFlow dataset was collected in-house across two physically distinct indoor environments: a Laboratory (7m x 7m) and a Classroom (5.5m x 9m), as shown in Fig. 6. This multi-domain approach is crucial for assessing a model's robustness, as performance can often degrade significantly when a model is deployed in an environment different from where it was trained~\cite{zhang2018crosssense}. Data was collected from 10 participants. During each trial, participants would enter or exit the monitored space following audible beeps timed 10 seconds apart. The UNIX timestamp of each beep was recorded, providing a precise ground truth for each `enter' and `exit' event.

The WiAR dataset~\cite{guo2019wiar} is a public benchmark for WiFi-based activity recognition. We use it as a challenging target domain to evaluate the transferability of the features learnt by our self-supervised encoder. For this task, we focus on classifying six distinct activities: `walk', `forward\_kick', `side\_kick', `sit\_down', `squat', and `bend'.

\begin{figure}[htbp]
  \centering
  \subfloat[Lab]{%
    \includegraphics[width=.9\linewidth]{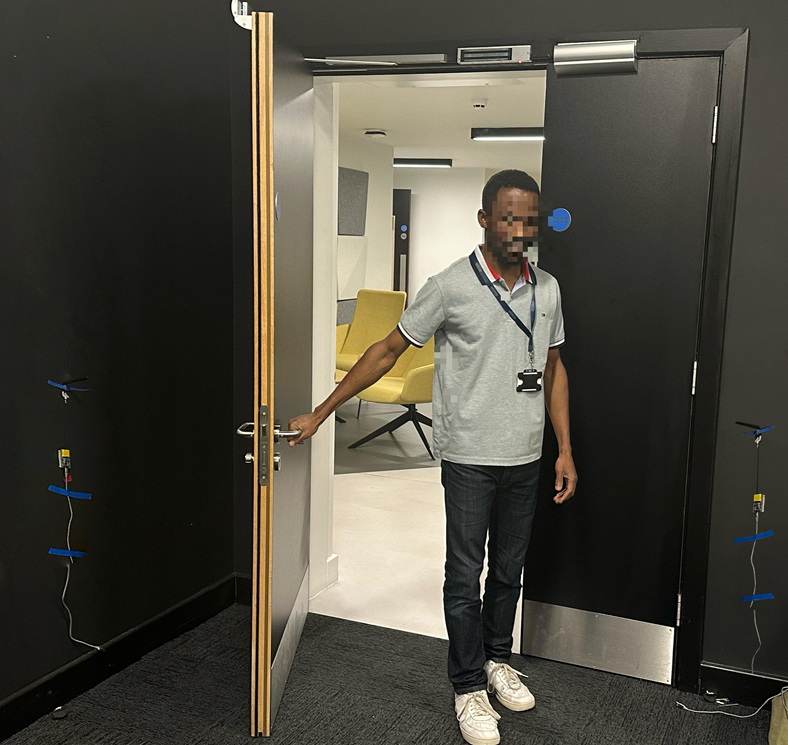}%
    \label{fig:6a}
  }\\[3mm]

  \subfloat[Classroom]{%
    \includegraphics[height=0.35\textheight]{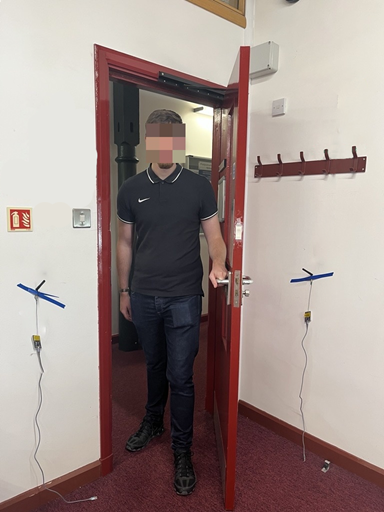}%
    \label{fig:6b}
  }

  \caption{The two environments used in our WiFlow experiments: (a) Lab and (b) Classroom.}
  \label{fig:stacked_images}
\end{figure}

\subsection{Evaluation Metrics}
To evaluate our framework, we employ a set of tailored metrics. For the underlying event classification, we report the overall Accuracy for general correctness and the Weighted F1-Score to provide a balanced measure that accounts for the inherent class imbalance between `enter', `exit', and `no-event' windows. The final, user-facing crowd counting performance is assessed primarily by the Mean Absolute Error (MAE), which represents the average error in the occupancy count over time. We also use the Root Mean Squared Error (RMSE) to more heavily penalize large, transient errors in the count. Finally, to quantify performance loss in cross-domain scenarios, we introduce a Generalisation Index (GI), calculated as the ratio of target-domain performance to source-domain performance for both accuracy and MAE. A GI score of 1.0 signifies perfect generalisation, whereas lower values indicate a performance drop when transferring to a new environment.

\section{Results and Discussion}
This section presents a comprehensive evaluation of our CSI-ResNet-A framework, validated across both our custom WiFlow dataset and the public WiAR benchmark. On our WiFlow dataset, we assess the full pipeline's effectiveness by reporting on both intermediate event classification and the final occupancy counting accuracy. For the public WiAR benchmark, we evaluate the transferability of our learnt features by focusing solely on the six-activity classification task.

\subsection{Performance on the WiFlow Dataset}

\begin{figure}[b!]
  \centering
  \subfloat[Event classification performance (Accuracy).]{%
    \includegraphics[width=\columnwidth]{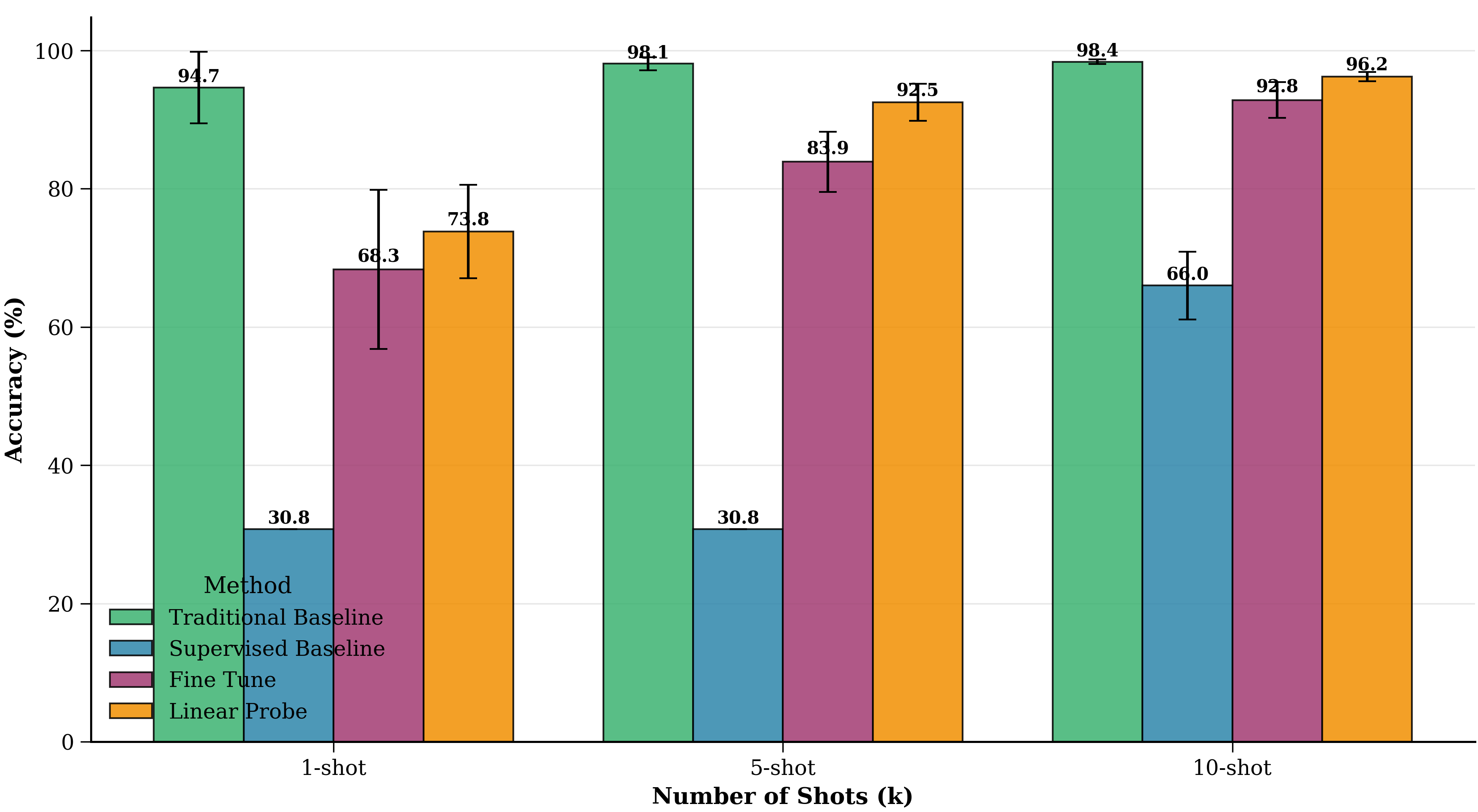}%
    \label{fig:7a}
  }\\[3mm]

  \subfloat[Final occupancy counting performance (MAE).]{%
    \includegraphics[width=\columnwidth]{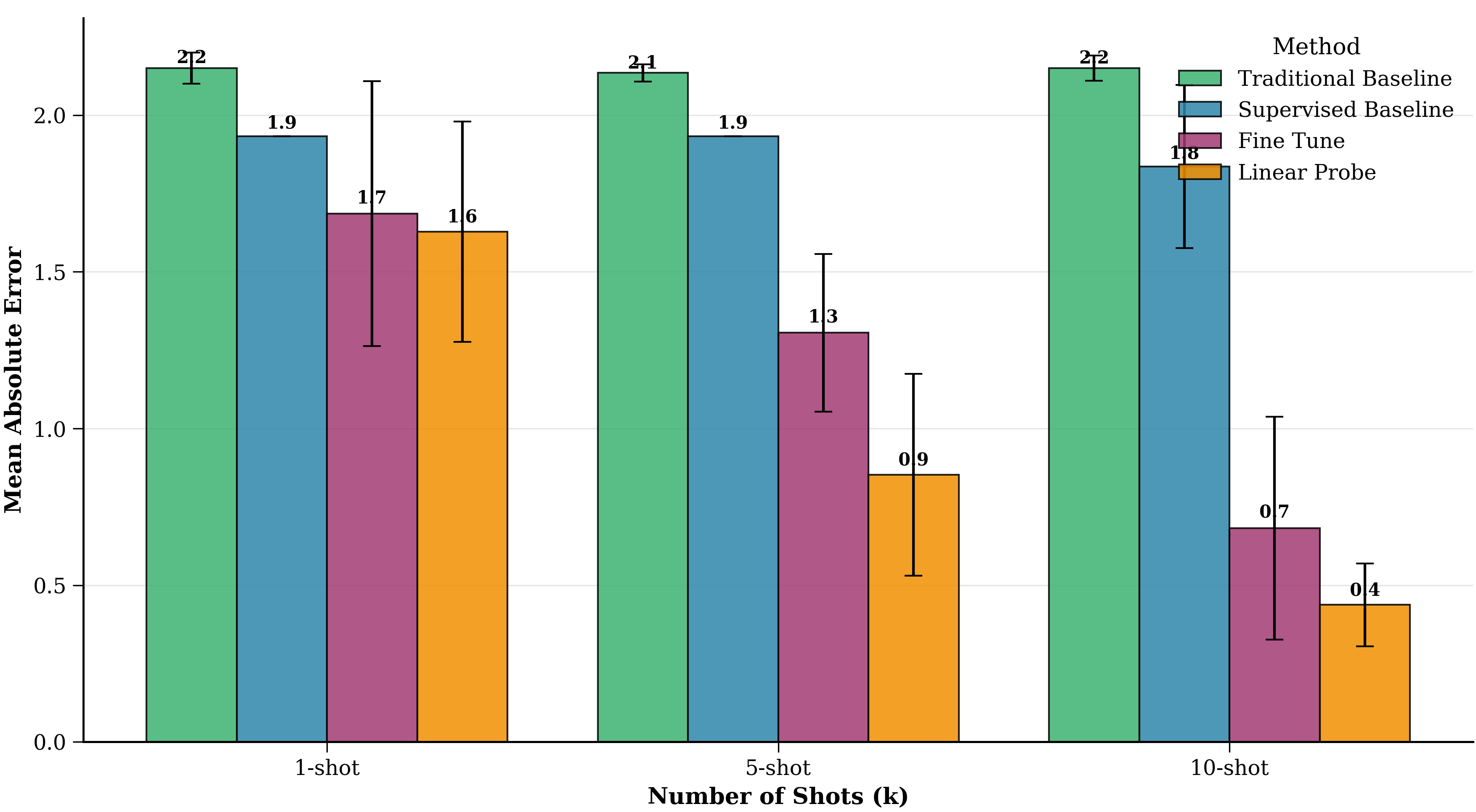}%
    \label{fig:7b}
  }

  \caption{Few-shot learning performance on the WiFlow dataset.}
  \label{fig:fig7}
\end{figure}

\subsubsection{Few-Shot Learning Evaluation}
We first evaluate our framework in a data-scarce, few-shot learning context, using k-shot values of 1, 5, and 10. The goal is to assess not only the classification performance but also how this performance translates to the final, practical task of occupancy counting.

\textbf{Event Classification:} The classification results, presented in Fig~\ref{fig:7a}, reveal a crucial insight into the limitations of different learning paradigms. The Supervised Baseline, trained from scratch, fails to learn effectively, achieving a stagnant accuracy of 30.8\% (F1-score of 14.5\% ±0.0) for both 1-shot and 5-shot scenarios. Conversely, the Traditional Baseline (Random Forest) demonstrates high performance in this isolated classification task, achieving an accuracy of 94.7\% (±5.19) with just a single labelled shot. Our unsupervised methods, Fine-Tune and Linear Probe, showcase the power of pre-training by starting with strong performance and steadily improving. The Linear Probe, which uses frozen features from the unsupervised encoder, becomes highly competitive, achieving a final accuracy of 96.2\% (±0.68) and an F1-score of 96.3\% (±0.67) at 10 shots. 

\textbf{Occupancy Counting:} The most critical results emerge when evaluating the final occupancy counting performance, shown in Fig~\ref{fig:7b}. Despite its exceptional classification accuracy, the Traditional Baseline completely fails at the counting task, yielding a high and stagnant cleaned MAE of approximately 2.15 across all k-shot values. This indicates a significant disconnect between per-window classification and time-series event tracking.

This is where our unsupervised approach proves its superiority. The Linear Probe method, leveraging the temporally coherent features from the pre-trained encoder, achieves a remarkable reduction in counting error. Its cleaned MAE improves dramatically from 1.63 at 1-shot to an excellent 0.44 at 10-shot. The Fine-Tune model shows a similar trend, with its cleaned MAE dropping to 0.68. This outcome is the central finding of our work: unsupervised pre-training learns representations that are not only accurate but also temporally stable, which is essential for reliable performance in real-world, stateful tasks like crowd counting.

\subsubsection{Zero-Shot Cross-Domain Generalisation}

\begin{figure}[t!]
  \centering
  \subfloat[Zero-shot classification accuracy.]{%
    \includegraphics[width=\columnwidth]{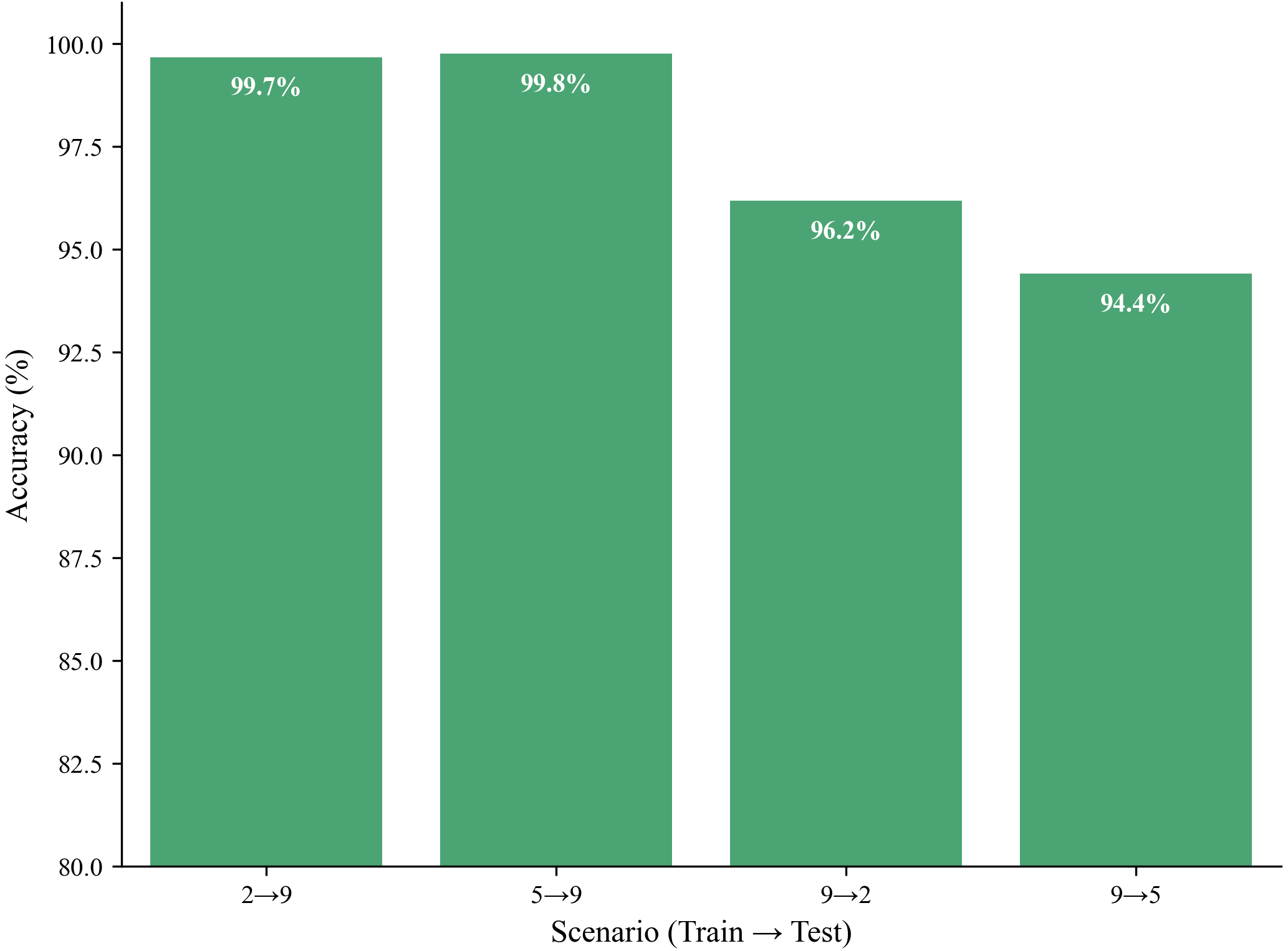}%
    \label{fig:8a}
  }

  \subfloat[Zero-shot occupancy counting MAE.]{%
    \includegraphics[width=\columnwidth]{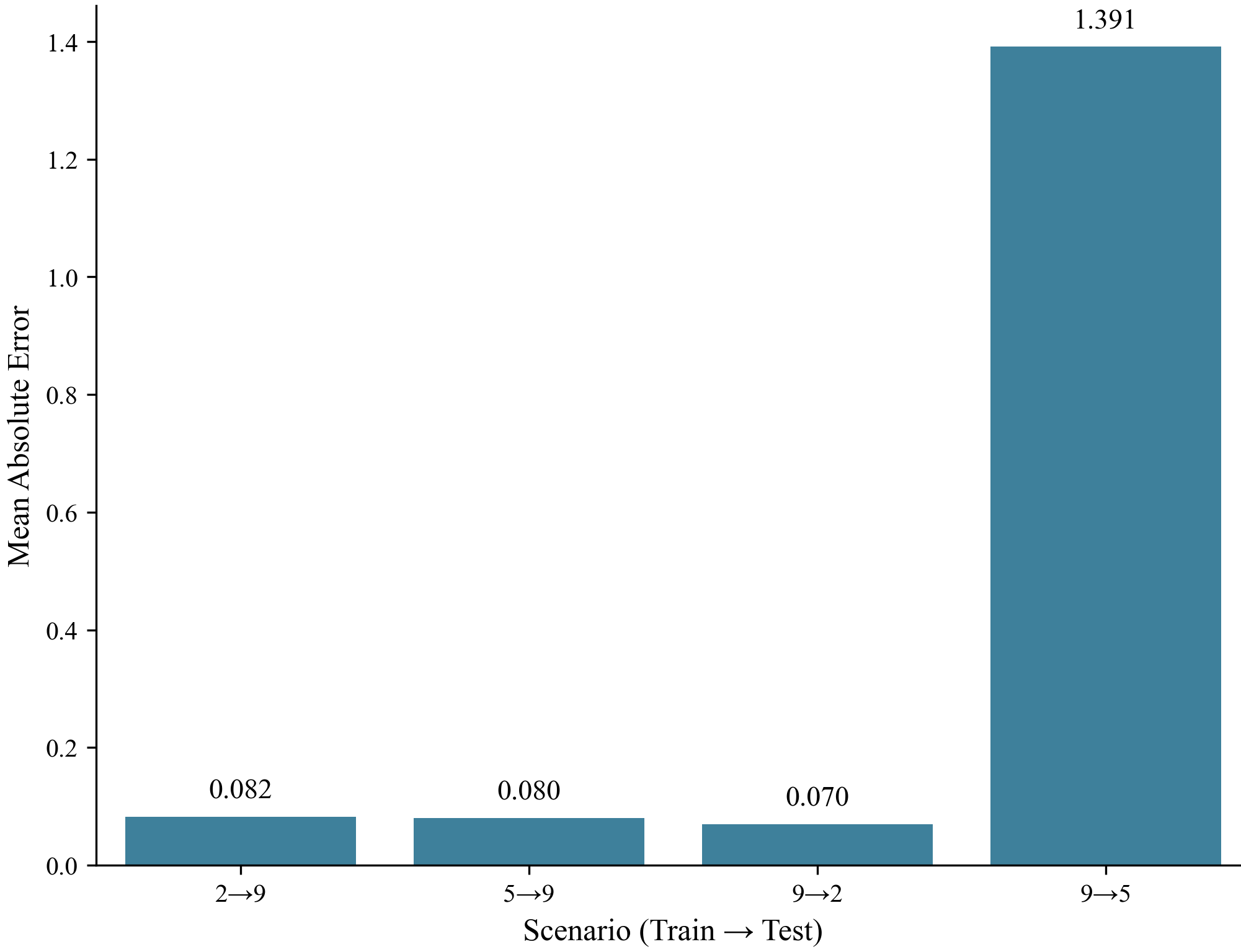}%
    \label{fig:8b}
  }

  \caption{Zero-shot cross-domain performance on the WiFlow dataset.}
  \label{fig:fig8}
\end{figure}

\begin{figure}[t!]
  \centering
  \subfloat[Generalisation Index for Accuracy.]{%
    \includegraphics[width=\columnwidth]{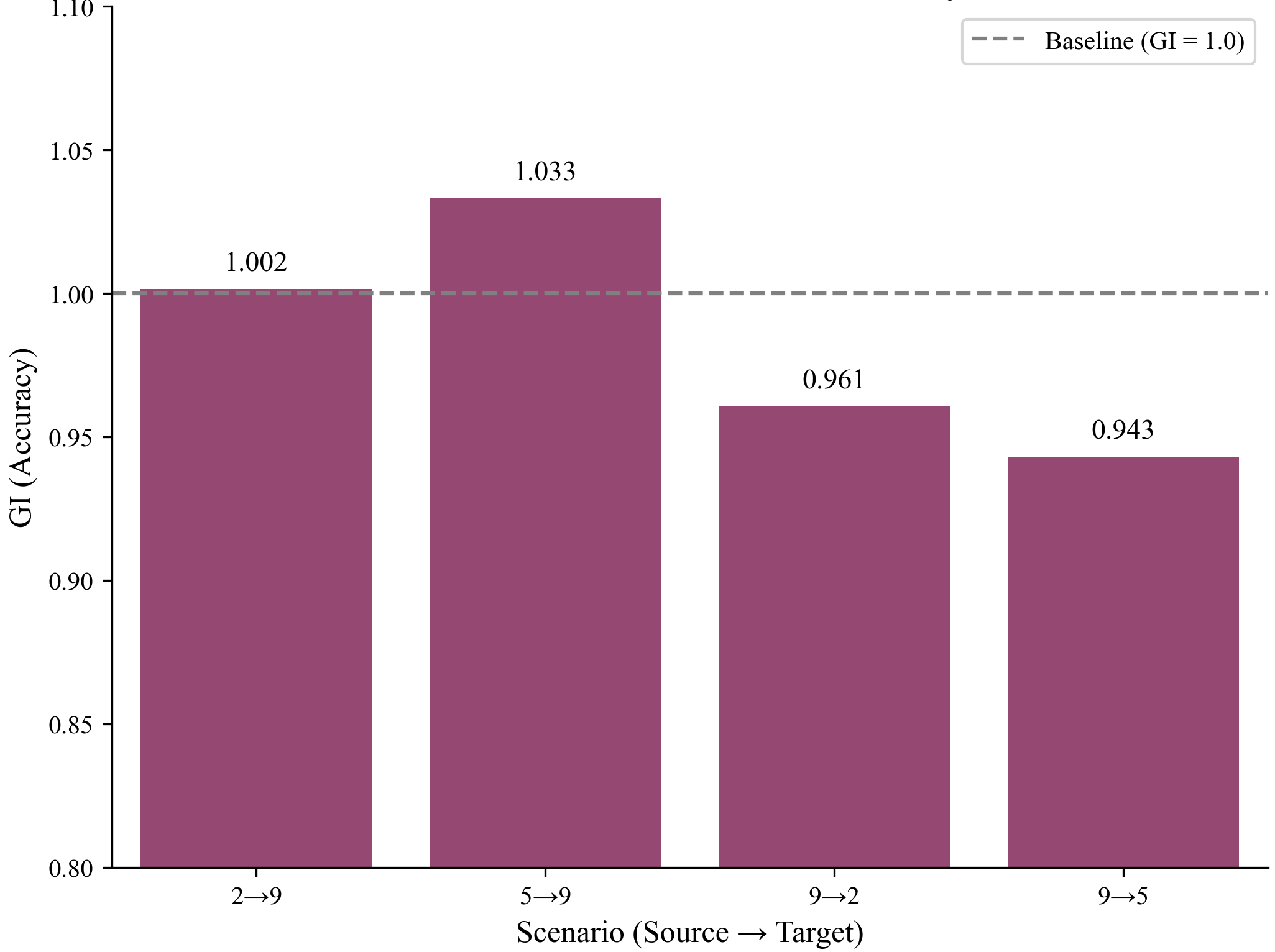}%
    \label{fig:9a}
  }\\[3mm]

  \subfloat[Generalisation Index for MAE (Higher is Better).]{%
    \includegraphics[width=\columnwidth]{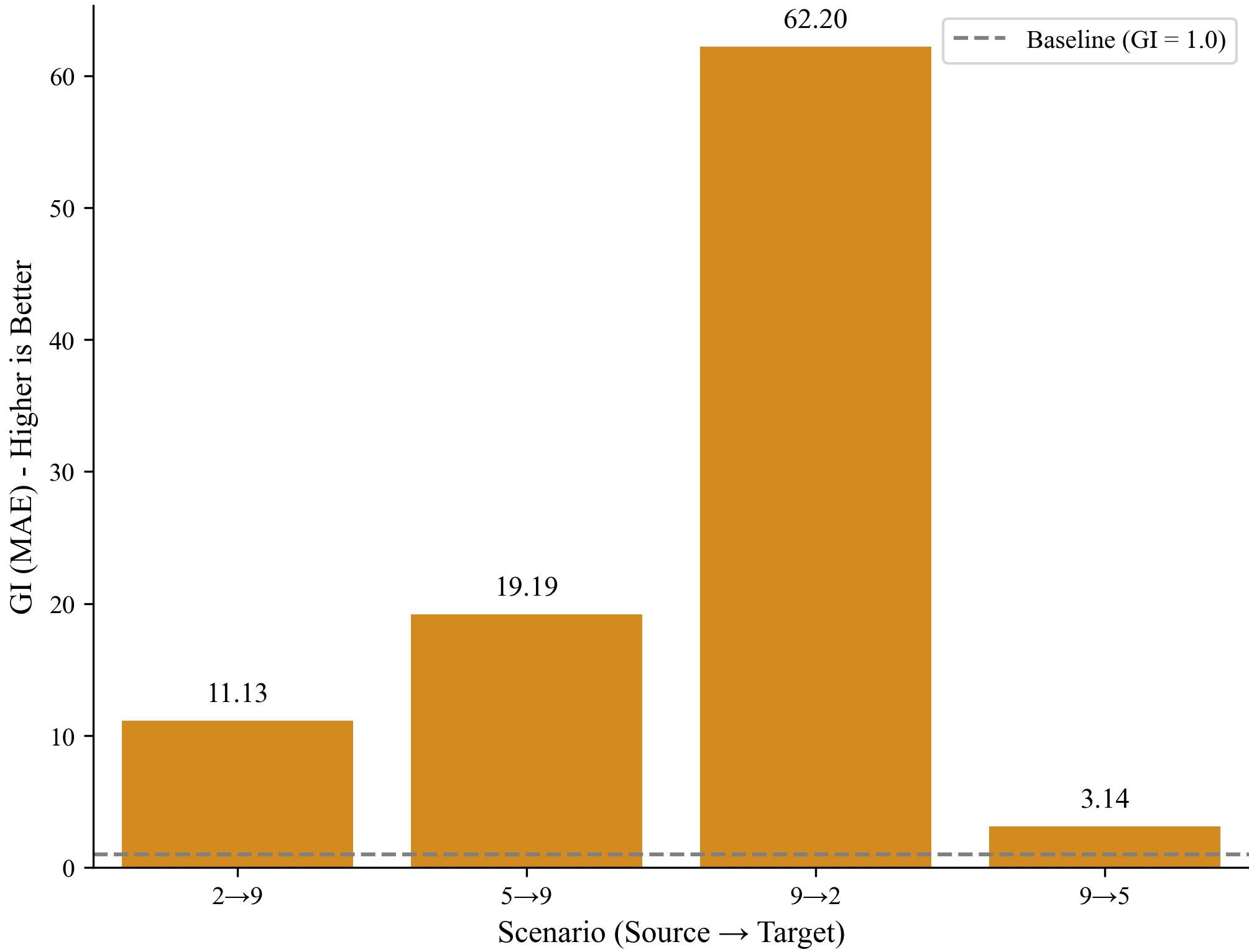}%
    \label{fig:9b}
  }

  \caption{Generalisation Index (GI) for accuracy and MAE on the WiFlow dataset.}
  \label{fig:fig9}
\end{figure}
A key measure of a robust feature representation is its ability to generalise to new, unseen environments. We evaluate this through a zero-shot cross-domain experiment where a model is trained on data from one environment (e.g., a 2-person household, denoted ``2") and tested directly on data from a different environment (e.g., a 9-person household, denoted ``9") without any fine-tuning.

The results, shown in Fig~\ref{fig:fig8}, demonstrate excellent Generalisation capabilities. In all scenarios, the classification accuracy remains remarkably high, exceeding 94\% (Fig~\ref{fig:8a}). For instance, when transferring from a 5-person to a 9-person household (5\textrightarrow9), the model achieves 99.8\% accuracy. This indicates that the features learnt during unsupervised pre-training are discriminative and robust across different environmental conditions and user densities.

However, the Mean Absolute Error for the final counting task (Fig~\ref{fig:8b}) reveals a more nuanced story. Although the model generalises exceptionally well in most cases, achieving MAEs as low as 0.07 (9\textrightarrow2), the transfer from the 9-person to the 5-person household (9\textrightarrow 5) results in a significantly higher MAE of 1.39. This suggests that while the features are still good enough for high classification accuracy, the subtle differences in this specific target domain impact the temporal consistency required for perfect counting.

To quantify this generalisation capability more formally, we use the Generalisation Index (GI), with results shown in Fig.~\ref{fig:fig9}. The GI for accuracy (Fig~\ref{fig:9a}) remains close to the ideal value of 1.0 in all scenarios, confirming the small drop in classification performance. The GI for MAE (Fig~\ref{fig:9b}) is more revealing. In scenarios like 9\textrightarrow2, the GI (MAE) is exceptionally high (62.2), indicating that the model performed significantly better on the unseen target domain than on the hold-out set of its source domain.

This counter-intuitive result stems from the differing complexity of the domains. The source domain contains more complex and varied event signatures, forcing the model to learn highly robust and discriminative features to achieve a reasonable source MAE. When this highly capable model is transferred to a less complex target domain (e.g., a 2-person household with more regular, rhythmic events), it finds the task comparatively simple. Its learnt robustness over-qualifies it for this new domain, resulting in an extremely low target MAE and, consequently, a very high GI (MAE). This demonstrates that our unsupervised pre-training learns universal, environment-agnostic features that are robust enough to handle complex scenarios and excel in simpler ones.

\subsubsection{Interpreting the Generalisation Index (GI)}
The strong performance of our metric, GI, is not incidental but rather a direct consequence of both our experimental design and the nature of the features learnt through self-supervision.

Our WiFlow dataset was intentionally designed with domains of varying complexity (e.g., different room layouts and numbers of occupants). This creates challenging transfer-learning scenarios. The exceptionally high GI (MAE) observed when transferring from a complex source domain to a simpler one is a key finding. For instance, in the 9\textrightarrow2 scenario, the model achieves a source MAE of 4.37 on its own complex domain but an exceptionally low target MAE of just 0.07 on the simpler, unseen domain. This results in a GI (MAE) of 62.2, quantitatively demonstrating that the contrastive pre-training on the more difficult domain compels the model to learn highly robust, discriminative features that are "over-qualified" for the simpler target environment. The model learns to isolate the fundamental signature of human movement from a noisy, complex background, and these learnt features are then highly effective when the background noise is reduced.

Furthermore, the quality of these learnt representations is strongly corroborated by their performance on the entirely separate WiAR benchmark. While the GI metric was not applied to this dataset, the fact that features pre-trained on WiFlow can achieve state-of-the-art accuracy on a different task (activity recognition) provides powerful external validation. This success demonstrates that our self-supervised pre-training has indeed captured genuinely universal, domain-invariant features of human-induced CSI perturbations. Consequently, it gives us high confidence that the strong GI scores observed on the WiFlow dataset are not a statistical artifact, but are instead a true measure of this powerful Generalisation capability.

\subsection{Performance on Public WiAR Benchmark}

\begin{figure*}[t!]
    \centering
    \includegraphics[width=\textwidth]{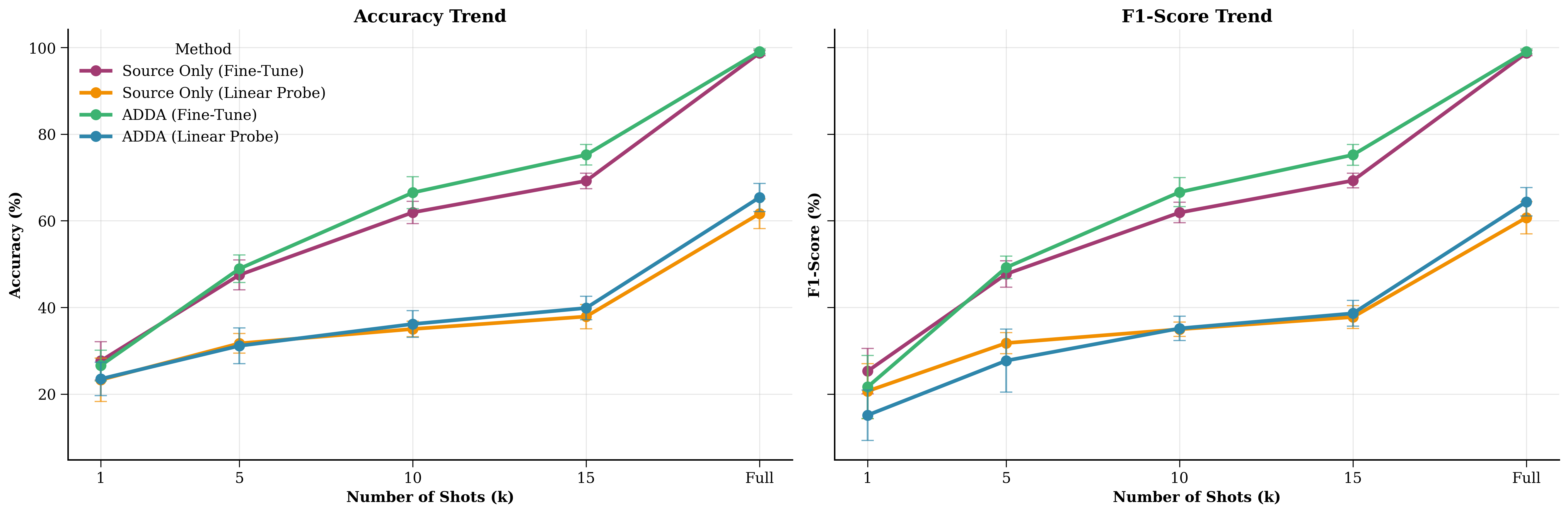}
    \caption{Accuracy and F1-Score on the WiAR benchmark. The plot compares four transfer learning strategies: Source Only (Fine-Tune), Source Only (Linear Probe), ADDA (Fine-Tune), and ADDA (Linear Probe).}
    \label{fig:fig10}
\end{figure*}
To validate the generalizability of our unsupervised pre-training, we benchmark our framework on the public WiAR dataset. We compare two deep learning transfer strategies: Source Only, where a model pre-trained on our WiFlow data is used directly on the target domain with no adaptation, and ADDA, which first uses unsupervised adversarial training to align the feature distributions of the source and target domains. These are evaluated against two robust benchmarks: the original state-of-the-art performance of 94.6\% reported by Guo et al.~\cite{guo2019wiar} and a strong Traditional Baseline (Random Forest on handcrafted features) that we implemented. Given the complexity of this six-class recognition task, we extended our evaluation to include 15-shot and full-dataset scenarios to better understand performance scaling.

The performance trends, shown in Fig~\ref{fig:fig10}, clearly illustrate the superiority of our proposed adaptation strategy. The two Linear Probe methods perform poorly, with the Source Only variant starting at 23.3\% accuracy (k=1) and plateauing at only 61.6\% on the full dataset. This indicates that without adaptation, a significant domain shift between WiFlow and WiAR limits performance. In contrast, the Traditional Baseline is surprisingly effective in low-data scenarios, achieving 63.5\% accuracy at just 1-shot and 88.1\% at 5-shots, significantly outperforming the deep learning models at these initial stages.

The impact of adaptation becomes clear as more data is available. The Source Only (Fine-Tune) method, while starting low, shows strong scaling, reaching an impressive 98.7\% accuracy on the full dataset. However, the most compelling result comes from the ADDA (Fine-Tune) method. Although it begins behind the traditional baseline, its superior learning rate is evident. At 10 shots, it achieves 66.5\% accuracy, and by 15 shots, it reaches 75.2\%, decisively closing the gap. This culminates in a final accuracy of 98.8\% and an F1-score of 98.8\% on the full dataset.

This result establishes a new state-of-the-art on the WiAR dataset. It demonstrates that while direct feature transfer is beneficial, the optimal strategy is to first use unsupervised domain adaptation to create domain-invariant features and then fine-tune them. This two-stage process allows our model to surpass not only the original benchmark (94.6\%) but also the strong traditional baseline (97.3\%), proving it is the most effective strategy for this task.

\subsubsection{Analysis of learnt Embeddings}
To qualitatively evaluate the features learnt, we analyse the structure of the embedding space for our best, worst, and baseline transfer learning models. We extract feature embeddings from the encoder, project them into a 2D space using t-SNE, and train a Graph Convolutional Network (GCN) on a k-NN graph of these embeddings to get a quantitative measure of their class separability.

\begin{figure}[htbp]
    \centering
    
    \begin{subfigure}[b]{0.48\columnwidth}
        \centering
        \includegraphics[width=\textwidth]{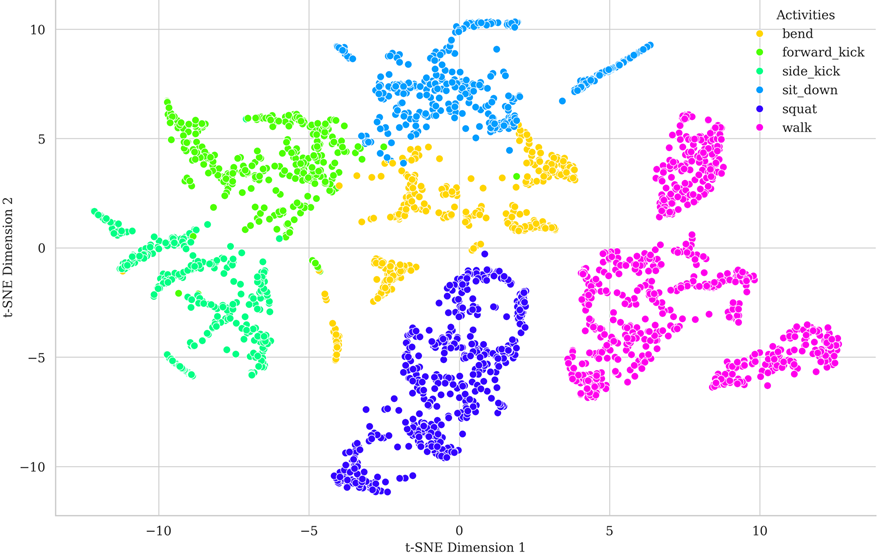}
        \caption{ADDA (Fine-Tune) - GCN Accuracy: 99.3\%.}
        \label{fig:11a}
    \end{subfigure}
    \hfill
    \begin{subfigure}[b]{0.48\columnwidth}
        \centering
        \includegraphics[width=\textwidth]{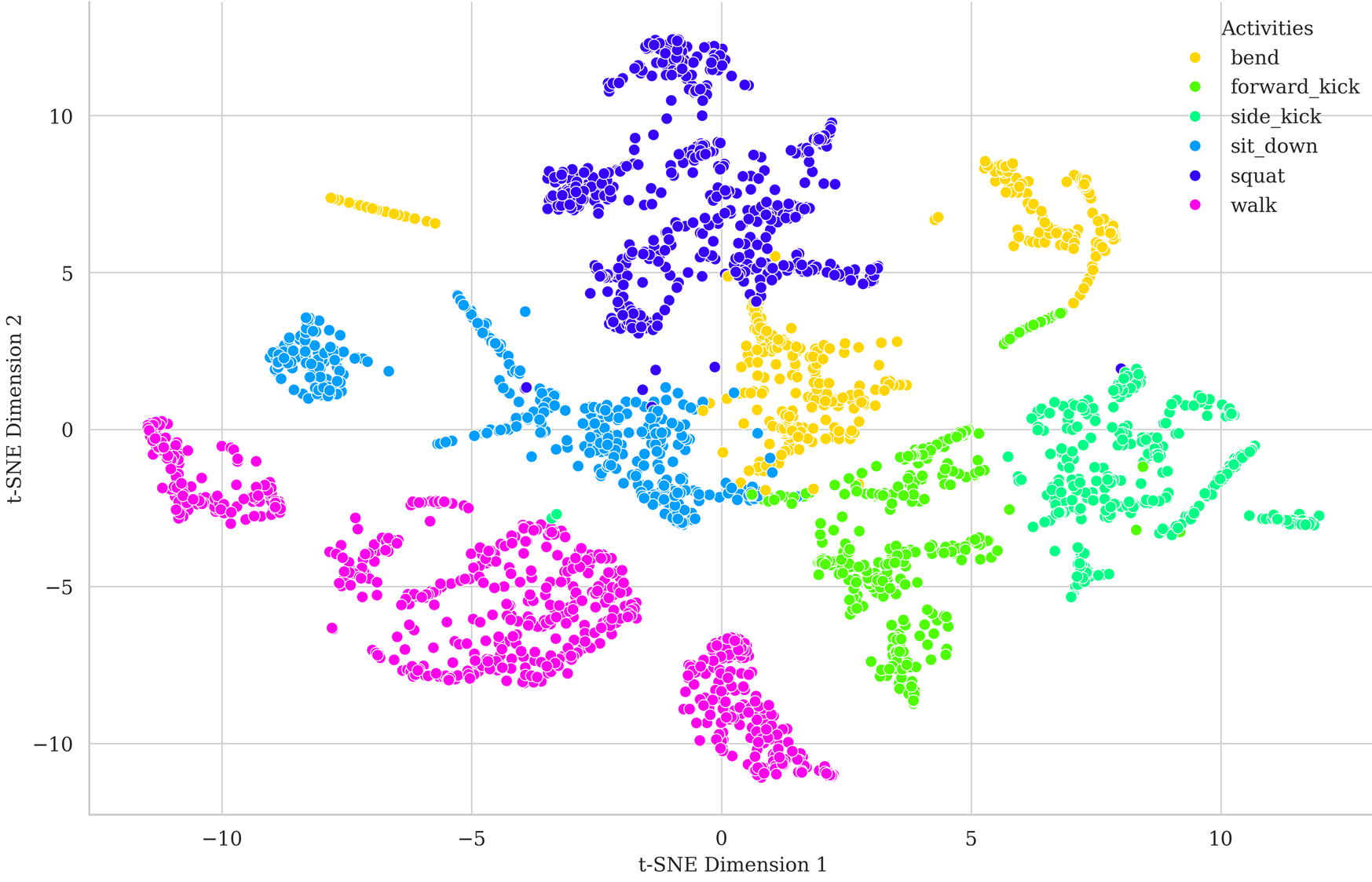}
        \caption{Source Only (Fine-Tune) - GCN Accuracy: 98.7\%.}
        \label{fig:11b}
    \end{subfigure}
    
    \begin{subfigure}[b]{0.48\columnwidth}
        \centering
        \includegraphics[width=\textwidth]{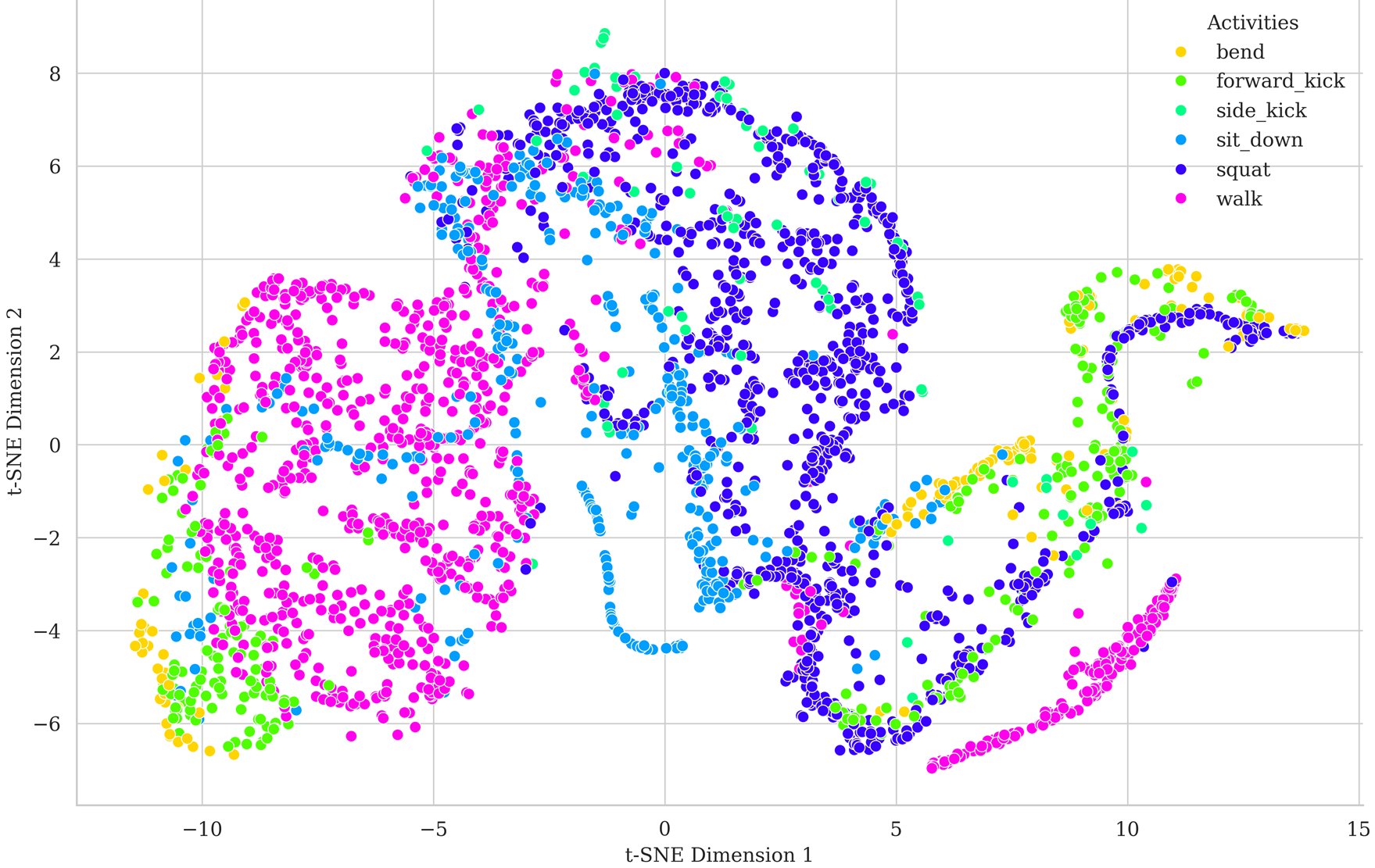}
        \caption{Source Only (Linear Probe, 1-shot) - GCN Accuracy: 45.4\%.}
        \label{fig:11c}
    \end{subfigure}
    
    \caption{t-SNE visualizations of feature distributions. The plots show the embeddings for (a) the best-performing ADDA model, (b) the Source Only fine-tuned model, and (c) the worst-performing 1-shot linear probe model.}
    \label{fig:11}
\end{figure}

The results, visualised in Fig~\ref{fig:11}, are striking. The bottom panel shows the embeddings from the 1-shot Source Only (Linear Probe) model. The classes are almost completely overlapping, and the GCN classifier achieves an accuracy of only 45.4\%, confirming that the raw, unadapted features are not separable in the low-data regime.

The top-right panel shows the Source Only (Fine-Tune) model. Here, the embeddings form distinct, well-separated clusters for each of the six activity classes. The GCN accuracy for this model is 98.7\%, demonstrating that fine-tuning adapts the pre-trained features effectively to the new domain.

The top-left panel shows the embeddings after applying ADDA (Fine-Tune). The class clusters are even tighter and more distinctly separated than in the Source Only model. This visual improvement is confirmed by the GCN classifier, which achieves a near-perfect accuracy of 99.3\%. This demonstrates that while fine-tuning alone produces high-quality features, the initial step of unsupervised domain adaptation creates a more robust and linearly separable feature space, leading to the best possible performance.

\subsubsection{Ablation Studies}
To quantify the contribution of each component in our proposed framework, we conduct a series of ablation studies on the WiAR dataset. We systematically evaluate the impact of our model's architectural components and the pre-training strategy. The complete results are presented in Table~\ref{tab:table2}.

We analyse the impact of the two primary architectural components: Squeeze-and-Excitation (SE) blocks and parameter-efficient adapter modules.

\textbf{SE Blocks:} The importance of the SE blocks is evident across all experiments. Their removal consistently degrades performance, confirming their value in learning discriminative CSI features. In the full fine-tuning scenario, disabling SE blocks leads to a minor accuracy drop from 99.67\% to 99.34\%. However, the effect is far more pronounced in the parameter-efficient setting, where the same change causes accuracy to fall by over 1\%, from 98.01\% to 96.84\%. This disparity highlights that when the number of trainable parameters is intentionally limited, the model's ability to dynamically recalibrate channel-wise feature responses becomes critical. The SE blocks force the network to focus on the most informative features, a benefit that is magnified when the model's learning capacity is constrained, a finding supported by their successful application in other time-series domains such as Human Activity Recognition~\cite{khan2025comparative}.

\textbf{Adapters:} The most significant finding of our study relates to the immense efficiency of parameter-efficient fine-tuning (PEFT) using adapters. Adapters are part of a growing family of PEFT methods designed to adapt large pre-trained models with minimal computational cost. While our work employs bottleneck adapters inserted within the transformer layers, other prominent techniques take different approaches. Low-Rank Adaptation (LoRA), for instance, injects trainable low-rank matrices into the layers of a frozen network~\cite{hu2022lora}. Other methods modify the input to the model, such as Prompt Tuning, which learns a small set of "soft prompts" that are prepended to the input sequence~\cite{lester2021power}, or Prefix Tuning, which adds learnable prefix vectors to the keys and values in the attention mechanism~\cite{li2021prefix}. A simpler approach, BitFit, involves fine-tuning only the bias terms of the network~\cite{zaken2021bitfit}.

During the fine-tuning process, the weights of the original encoder, which contains the vast majority of the model's parameters, are kept frozen. Only the parameters of these newly-added, lightweight adapter modules are updated.

Our findings strongly validate the use of adapters in this context. As shown in Table~\ref{tab:table2}, a full fine-tuning of the encoder requires updating 1,092,806 parameters to achieve a peak accuracy of 99.67\%. In stark contrast, our adapter-based method achieves a highly competitive accuracy of 98.84\% while updating only 30,438 parameters. This represents a staggering 97.2\% reduction in trainable weights for a performance drop of less than 1\%. By isolating the task-specific adaptations to a small set of parameters, we effectively steer the model’s behavior without incurring the significant computational and memory costs of full fine-tuning. This aligns with the broader trend in the field, demonstrating that PEFT methods can match the performance of full fine-tuning at a fraction of the cost, a principle now well-established in computer vision~\cite{li2024adapter}.

\begin{table}[t!]
\centering
\caption{Exhaustive Ablation Study Results on the WiAR Dataset}
\label{tab:table2}
\footnotesize
\setlength{\tabcolsep}{4pt} 
\begin{tabular*}{\columnwidth}{@{\extracolsep{\fill}} l l c c r}
\toprule
\textbf{Method} & \textbf{Augmentation} & \textbf{SE} & \textbf{Accuracy (\%)} & \shortstack{\textbf{Trainable}\\\textbf{Params}} \\ 
\midrule
Full-FT & Yes & Yes & \textbf{99.67} & 1,092,806 \\
Full-FT & Yes & No & 99.34 & 1,071,302 \\ 
\midrule
Adapter-FT & Yes & Yes & 98.01 & 30,438 \\
Adapter-FT & Yes & No & 96.84 & 30,438 \\
Adapter-FT & No & Yes & \textbf{98.84} & \textbf{30,438} \\ 
\midrule
Linear Probe & Yes & Yes & \textbf{61.30} & \textbf{774} \\
Linear Probe & Yes & No & 55.65 & 774 \\ 
\bottomrule
\end{tabular*}
\end{table}

\section{Conclusion}
This paper successfully addresses the critical domain shift problem in WiFi-based crowd-counting by introducing a novel two-stage framework built upon a CSI-ResNet-A architecture. Our approach leverages self-supervised contrastive learning to build a robust feature encoder and uses highly efficient adapter-based fine-tuning for adaptation to new environments.

The framework's effectiveness is demonstrated through a series of key findings. On our proprietary WiFlow dataset, our unsupervised approach excels in a data-scarce, 10-shot learning scenario, achieving a best-in-class event classification accuracy of 96.3\% with a linear probe and a final Mean Absolute Error (MAE) of just 0.44 for the occupancy count. This significantly outperforms supervised baselines, which fail to learn effectively. The model's robustness is further proven in zero-shot cross-domain tests, where it achieves a near-perfect Generalisation Index (GI) for accuracy, indicating minimal performance loss when transferring to new environments.

The generalisability of our learnt features is validated on the public WiAR benchmark, where our framework sets a new state-of-the-art classification accuracy of 98.8\%. Our ablation studies underscore the architectural strengths of our model. We confirm that Squeeze-and-Excitation (SE) blocks are critical for performance, particularly in parameter-constrained settings. Most significantly, we highlight the efficiency of our adapter-based fine-tuning, which achieves performance within 1\% of a full fine-tune (98.84\% vs. 99.67\%) while training 97.2\% fewer parameters.

Although our framework provides a robust solution for cross-domain crowd-counting, its success opens several promising avenues for future research. The current model is designed to handle discrete, single-person events; a key next step would be to extend it to more complex scenarios. This could involve reformulating the classification task to detect simultaneous multi-person entries and exits, or developing a more sophisticated state-tracking logic capable of interpreting rapid, overlapping event sequences.

Another critical direction is real-time deployment on edge devices. Although our adapter-based fine-tuning is highly efficient, the ResNet backbone may still be too demanding for low-power microcontrollers. Future work could explore model compression techniques such as pruning and quantisation, or use knowledge distillation to train a smaller, faster "student" model that retains the performance of our larger architecture, making on-premise, real-time inference practical.

Furthermore, the quality of our learnt representations could be enhanced by exploring more advanced self-supervised learning (SSL) techniques. Although our contrastive approach proved effective, methods such as Masked Auto-Encoding (MAE), which involve reconstructing masked portions of the input signal, could compel the model to learn even more nuanced and holistic features of the underlying CSI dynamics.

Finally, to improve robustness in exceptionally noisy RF environments, multi-modal sensor fusion presents a compelling research direction. Integrating our CSI-based system with other low-cost, privacy-preserving sensors, such as Passive Infrared (PIR) detectors or mmWave radar, could create a more resilient system. An attention-based fusion mechanism could be developed to allow the model to dynamically weigh the importance of each modality, learning to trust the most reliable data source at any given moment.

\bibliographystyle{IEEEtran}
\bibliography{Bibliography}

\end{document}